%% file: main_arxiv.tex
\documentclass[letterpaper]{article} 
\usepackage{aaai2026}  
\usepackage{times}  
\usepackage{helvet}  
\usepackage{courier}  
\usepackage[hyphens]{url}  
\usepackage{graphicx} 
\usepackage{natbib}  
\usepackage{caption} 
\usepackage{algorithm}
\usepackage{algorithmic}

\usepackage{bibentry}
\usepackage{multirow}
\usepackage{subcaption}
\usepackage{amsfonts}
\usepackage[inkscapelatex=false]{svg}

\usepackage{amsmath}
\usepackage{tabularx}
\usepackage{enumitem}
\usepackage{bibentry}

\newcommand{\LL}{{\mathcal{L}}}

\newcommand{\V}{{\mathcal{V}}}

\newcommand{\X}{{\mathbf{X}}}
\newcommand{\A}{{\mathbf{A}}}

\newcommand{\M}{{\mathbf{M}}}

\usepackage{newfloat}
\usepackage{listings}
\DeclareCaptionStyle{ruled}{labelfont=normalfont,labelsep=colon,strut=off} 
\floatstyle{ruled}
\newfloat{listing}{tb}{lst}{}
\floatname{listing}{Listing}

\title{Towards Multiple Missing Values-resistant Unsupervised Graph Anomaly Detection}

\author {
    Jiazhen Chen\textsuperscript{\rm 1}\equalcontrib,
    Xiuqin Liang\textsuperscript{\rm 2}\equalcontrib,
    Sichao Fu\textsuperscript{\rm 3, \rm 4}\thanks{Corresponding author.},
    Zheng Ma\textsuperscript{\rm 5},
    Weihua Ou\textsuperscript{\rm 6}
}
\affiliations {
    \textsuperscript{\rm 1}Department of Statistics and Actuarial Science, University of Waterloo\\
    \textsuperscript{\rm 2}Data Science Center of Excellence, Deloitte Consulting Beijing\\
    \textsuperscript{\rm 3}School of Electronic Information and Communications, Huazhong University of Science and Technology\\
    \textsuperscript{\rm 4} Text Computing and Cognitive Intelligence Ministry of Education Engineering Research Center, Guizhou University
    \textsuperscript{\rm 5}Cheriton School of Computer Science, University of Waterloo\\
    \textsuperscript{\rm 6}School of Big Data and Computer Science, Guizhou Normal University\\
    j385chen@uwaterloo.ca, pliang@deloittecn.com.cn, fusichao$\_$upc@163.com, z43ma@uwaterloo.ca, ouweihua@gznu.edu.cn
}

\begin{document}

\maketitle

\begin{abstract}

\input{abstract}

\end{abstract}


\section{Introduction}

\input{introduction_v1}

\section{Methodology}

\input{methodology_v1}

\section{Experiment}

\input{experiment_v1}

\section{Conclusion}

\input{conclusion_short}

\section{Appendix}

\subsection{Baselines and Evaluation Metric}

We benchmark five state-of-the-art unsupervised graph anomaly detection methods: 1) CoLA~\cite{liu2022cola}, which employs a local–global contrastive learning approach and trains a discriminator to identify inconsistencies between a node and its neighborhood; 2) PREM~\cite{pan2023prem}, which detects anomalies by directly measuring ego–neighbor similarity without message passing; 3) GRADATE~\cite{duan2023gradate}, which utilizes multi-scale contrastive learning with dual graph views to capture both local and global irregularities; 4) ADA-GAD~\cite{he2024ada}, which adopts a two-stage denoise-then-detect pipeline with anomaly-aware regularization; and 5) DiffGAD~\cite{li202diffgad}, which leverages diffusion models to score anomalies in the latent space.

Since none of these detectors natively handles graphs with missing nodes or edges, we adopt a two‐stage “impute‐then‐detect” strategy. First, we complete the graphs using one of five general‐purpose imputers—mean filling (column‐wise averages), MissForest~\cite{stekhoven2011missforest}, GAIN~\cite{yoon2018gain}, GAE~\cite{kipf2016variational}, or ASD‐VAE~\cite{jiang2024incomplete}. The first three methods ignore graph topology and impute only node features, whereas GAE and ASD‐VAE leverage the observed topology to guide feature reconstruction. Second, we apply each of the five detectors to each of the five imputed graphs, yielding 25 impute‐then‐detect baseline combinations.

For evaluation, we use the Area Under the Receiver Operating Characteristic curve (AUROC) as our primary evaluation metric. AUROC provides a threshold-independent measure of the model’s ability to distinguish between normal and anomalous nodes, and is widely adopted for anomaly detection due to its robustness to class imbalance. 

\subsection{Model Configuration}

All experiments are conducted on a single NVIDIA A100 GPU (40 GB memory) using PyTorch 2.2 and PyG 2.5. We train using the Adam optimizer, with the learning rate selected from the set $\{0.0001, 0.001, 0.0005\}$ according to the speed of training loss convergence. Both initial training and fine-tuning phases are run for 100 epochs each. The latent dimension $d_z$ is set to 256 for most datasets, except for Flickr and ACM, where it is reduced to 128 due to memory constraints. We use a two-layer GCN encoder in all experiments. The Sinkhorn divergence uses regularization $\epsilon = 0.1$ with 1000 iterations. For pseudo-anomaly generation, we sample $M=0.1n$ latent vectors, corresponding to 10\% of the training nodes. Loss component weights ($\alpha$, $\lambda$) are selected from $\{0.1, 0.01, 0.001, 0.0001\}$ based on the training loss, and we fix them at $\alpha=0.01$ and $\lambda=0.001$ for all datasets.

\subsection{Datasets}

 We perform experiments on seven real-world attributed graphs spanning diverse application domains, including Cora~\cite{yang2016revisiting}, Citeseer~\cite{yang2016revisiting}, Flickr~\cite{zeng2019graphsaint}, and ACM~\cite{tang2008acm}, Disney~\cite{muller2013disney}, Books~\cite{sanchez2013books}, and Reddit~\cite{kumar2019reddit}. Table~\ref{tab:dataset-stats} summarizes each graph’s number of nodes, number of edges, feature dimensionality, average node degree, number of anomalous nodes and anomaly ratio. 

\subsection{Pseudo code}

See Algorithm~\ref{Algorithm}

\input{pseudo_code}

\section*{Acknowledgements}

This work was supported in part by the National Natural Science Foundation of China under Grant 62262005, in part by the High-level Innovative Talents in Guizhou Province under Grant GCC[2023]033, in part by the Open Project of the Text Computing and Cognitive Intelligence Ministry of Education Engineering Research Center under Grant TCCI250208.

\bibliography{aaai2026}

\end{document}

%% file: abstract.tex
Unsupervised graph anomaly detection (GAD) has received increasing attention in recent years, which aims to identify data anomalous patterns utilizing only unlabeled node information from graph-structured data. However, prevailing unsupervised GAD methods typically presuppose complete node attributes and structure information, a condition hardly satisfied in real-world scenarios owing to privacy, collection errors or dynamic node arrivals. Existing standard imputation schemes risk ``repairing'' rare anomalous nodes so that they appear normal, thereby introducing imputation bias into the detection process. In addition, when both node attributes and edges are missing simultaneously, estimation errors in one view can contaminate the other, causing cross-view interference that further undermines the detection performance. To overcome these challenges, we propose M$^2$V-UGAD, a multiple missing values–resistant unsupervised GAD framework on incomplete graphs. Specifically, a dual-pathway encoder is first proposed to independently reconstruct missing node attributes and graph structure, thereby preventing errors in one view from propagating to the other. The two pathways are then fused and regularized in a joint latent space so that normals occupy a compact inner manifold while anomalies reside on an outer shell. Lastly, to mitigate imputation bias, we sample latent codes just outside the normal region and decode them into realistic node features and subgraphs, providing hard negative examples that sharpen the decision boundary. Experiments on seven public benchmarks demonstrate that M$^2$V-UGAD consistently outperforms existing unsupervised GAD methods across varying missing rates.

%% file: introduction_v1.tex
Graph-structured data is pervasive across various real-world domains, including social networks, financial transaction systems, e-commerce platforms, and biological interaction networks~\cite{chaudhary2019anomaly,huang2022dgraph,zhang2022efraudcom,chen2025semi,yu2024towards,yu2023sparse}. While these graph-structured data are adept at capturing rich and complex relationships among entities, they are frequently contaminated by anomalous patterns. For example, collusive users may inflate or deflate item ratings to manipulate ranking algorithms; whereas in social-media graphs, spam accounts frequently coalesce into unusually dense connectivity clusters~\cite{fakhraei2015collective,zhang2020gcn}. Graph anomaly detection (GAD) seeks to isolate these irregular nodes by identifying attribute profiles or structural patterns that diverge markedly from the dominant graph distribution. Nevertheless, the effectiveness of GAD is often hindered by extreme class imbalance and the scarcity of reliable labels: true anomalies are vastly outnumbered by normal nodes, and acquiring ground-truth annotations is both costly and often impractical~\cite{ma2021comprehensive}. These challenges have propelled recent research toward unsupervised GAD, where detectors must discern abnormality solely from the unlabeled graph-structured data.

Unsupervised GAD seeks to effectively spot nodes or substructures that violate the graph’s intrinsic attributes or topological regularities, operating on the premise that anomalies stand apart from normal patterns. Existing work falls largely into two families: reconstruction-based approaches and contrastive-learning approaches. Reconstruction-based methods train graph autoencoders to recover node attributes or structure and detect anomalies by measuring reconstruction errors~\cite{ding2019graphautoencoder,luo2022comga,he2024ada,li202diffgad}. For instance, ADA-GAD~\cite{he2024ada} pretrains multi-level graph autoencoders on spectral anomaly-denoised augmented graphs, then reconstructs the original graph and uses the resulting attribute and structure reconstruction errors to identify anomalies. DiffGAD~\cite{li202diffgad} leverages latent diffusion models to distill discriminative and common information in node representations, then reconstructs the original graph and uses reconstruction errors in the latent space to detect anomalies.

On the other hand, contrastive learning-based methods distinguish anomalies by learning to separate normal and abnormal patterns through agreement or disagreement between node representations and their local neighborhoods~\cite{liu2022cola,pan2023prem,jin2021anemone,duan2023gradate,chen2024towards}. For example, CoLA~\cite{liu2022cola} formulates anomaly detection as a node-versus-subgraph contrastive task, training a GNN to distinguish normal from abnormal nodes according to the agreement between node representations and their local subgraphs. PREM~\cite{pan2023prem} pre-aggregates ego-neighbor features with a single anonymized message-passing step, then uses a simple contrastive matching network to efficiently score anomalies according to ego-neighbor representation similarity.

Despite their efficacy, existing unsupervised GAD methods implicitly assume a complete input graph-structured data, i.e., one in which both the node feature matrix is complete and the adjacency matrix captures the full topology. Nevertheless, both the node attributes and edge connectivity are often missing in practice: node attributes may be redacted for privacy or lost in acquisition; node edges may be missing when new nodes arrive or when interactions go unrecorded \cite{xia2025incomplete}. A typical remedy is to perform data imputation before anomaly detection. Off-the-shelf imputers like Mean, MissForest \cite{stekhoven2011missforest} or GAIN \cite{yoon2018gain} treat each node in isolation, thus ignoring the relational cues that become critical under severe missingness. More recent graph‐aware imputers exploit the observed topology to guide node attribute reconstruction: some match the distributions of node attribute and structure embeddings~\cite{jiang2024incomplete,chen2020learning}, others refine structure-derived embeddings iteratively to impute missing features~\cite{tu2022initializing}, or restrict message passing to observed attributes only~\cite{jiang2020incomplete}. Yet these techniques hinge on reciprocal complementation between the node attribute and structural views; when either view is incomplete, noise in one bleeds into the other. Recent studies reveal that this cross-view interference amplifies reconstruction errors and ultimately degrades the resulting node representations~\cite{huo2023t2gnn,fu2023towards}. Some approaches mitigate this by decoupling attribute imputation from structure enhancement before alignment \cite{huo2023t2gnn,yuan2024mds}. However, they heavily rely on partial label supervision for node classification and are therefore unsuitable for a fully unsupervised anomaly detection task.

Aside from the mutual interference arising from dual feature–structure incompleteness, another critical challenge is the imputation bias issue, which arises due to the rarity of anomalies. In a conventional two-stage (impute-then-detect) pipeline, the imputer is trained almost exclusively on nodes that exhibit normal behaviour. As a result, it fills missing attributes or absent edges with prototypical normal values during inference. This process suppresses the distinctive patterns that characterise anomalies and pulls them toward the distribution of normal nodes, and consequently masks them from the detector. Recent empirical studies demonstrate that the more faithfully an imputer reproduces normal patterns, particularly when the proportion of missing data is low, the more pronounced this bias becomes. Ultimately, it leads to degradation in anomaly-detection performance~\cite{xiao2024unsupervised,zemicheal2019anomaly}.

In this work, we propose an unsupervised anomaly detection framework robust to graphs with incomplete node attributes and topology, termed M$^2$V-UGAD. The framework is specifically designed to tackle cross-view interference and imputation bias under anomaly scarcity arising in the incomplete-graph setting. To correctly impute data while preventing errors from propagating between the attribute and structural views, we introduce a dual-pathway encoder that independently imputes node features via an MLP and restores missing edges through deterministic Personalized PageRank diffusion. To allow the reconstructed two views to complement one another, we fuse them in a shared latent space using a lightweight GCN. A Sinkhorn divergence further regularizes the embeddings toward a truncated hyperspherical Gaussian, ensuring a compact normal region and clear separation of anomalies. To preserve intricate attribute–structure semantics and avoid collapse of the normal region, we append a recovery decoder that reconstructs node features from the latent embeddings, enforcing semantic consistency. To counteract imputation bias under anomaly scarcity, we synthesize pseudo-anomalies by sampling latent codes just beyond the normal region, and decoding them into realistic features and an internally connected subgraph. Thus providing challenging and plausible contrastive examples. Finally, to reinforce a clear boundary between normal nodes and anomalies, we fine-tune with a mixed Sinkhorn loss, ensuring normals remain tightly clustered while pseudo-anomalies are aligned to the outer ring.

The main contributions are summarized as follows:

\begin{itemize}

    \item \textbf{Novel Problem:} To the best of our knowledge, we are the first to investigate and address the problem of unsupervised GAD in the setting where both node attributes and the underlying topology are incomplete.
    
    \item \textbf{Dual-pathway Imputation and Hyperspherical Fusion:} To avoid error propagation while exploiting complementary signals, we disentangle attribute and topology imputation via a dual-pathway module and fuse the two restored views under a truncated hyperspherical Gaussian prior. A reconstruction constraint is enforced to preserve fine-grained attribute–structure semantics.
    
    \item \textbf{Graph Pseudo-anomaly Generation:} To counteract imputation bias and fortify the anomaly frontier, we sample latent codes from the ring-shaped shell beyond the normal manifold, decode them into realistic pseudo-anomaly nodes and a self-contained subgraph, and enforce their separation via mixed Sinkhorn divergence.
    
    \item \textbf{Extensive Empirical Validation:} Experiments on seven public benchmarks show the superior performance of M$^2$V-UGAD over existing unsupervised GAD methods across varying missing rates.

\end{itemize}

%% file: methodology_v1.tex
\begin{figure*}[t]
    \centering
    \includegraphics[width=\linewidth]{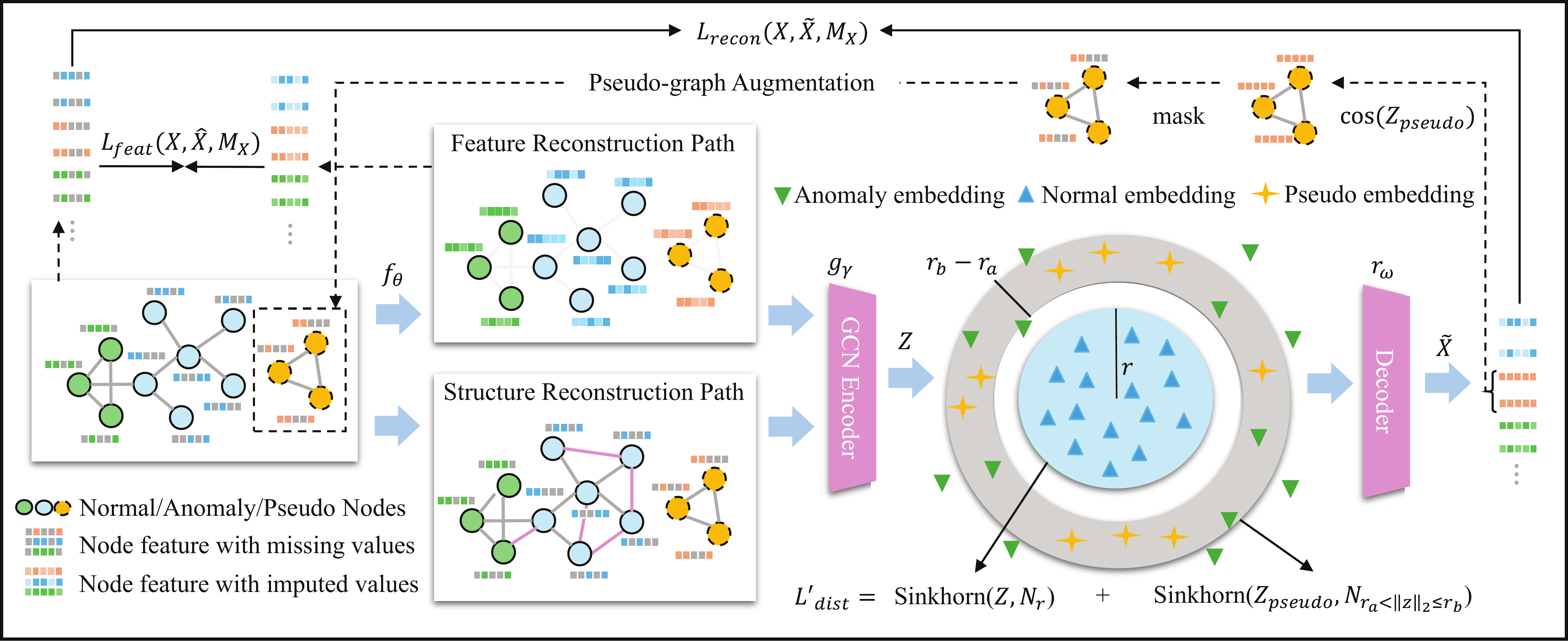}
    \caption{A diagram illustrating the proposed M$^2$V-UGAD framework.}
    \label{fig:architecture}
\end{figure*}

\subsection{Overview}

This section gives an overview of the M$^2$V-UGAD framework (see Figure~\ref{fig:architecture}), which tackles unsupervised GAD with incomplete attributes and topology. To handle missing features and edges without mutual interference, a dual-pathway imputation module independently reconstructs node attributes via an MLP encoder and restores missing edges via deterministic Personalized PageRank diffusion (Section~\ref{sec:dual-pathway}). These two reconstructions are projected into a shared latent space, where a Sinkhorn divergence aligns the empirical distribution to a truncated Gaussian on a hypersphere, and a recovery decoder reconstructs node attributes to preserve attribute–structure semantics (Section~\ref{sec:joint-latent}). Finally, to mitigate imputation bias from anomaly scarcity, the model is fine-tuned on an augmented dataset that includes pseudo-anomalies sampled from the ring-shaped shell beyond the normal latent sphere (Section~\ref{sec:pseudo-anomaly}). The full training pipeline and anomaly scoring are detailed in Section~\ref{sec:training}.

\subsection{Problem Formulation}
\label{sec:problem}

Consider an undirected attributed graph \(\mathcal{G}=(\mathcal{V},\mathcal{E},\mathbf{X})\), where \(\mathcal{V}=\{v_1,\dots ,v_n\}\) denotes the set of nodes, \(\mathcal{E}\subseteq\mathcal{V}\times\mathcal{V}\) denotes the set of edges, and \(\mathbf{X}=[\mathbf{x}_1;\dots ;\mathbf{x}_n]\in\mathbb{R}^{n\times d}\) is the node attribute matrix, with \(\mathbf{x}_i\in\mathbb{R}^{d}\) representing the attributes of node \(v_i\). The graph topology is encoded by a binary symmetric adjacency matrix \(\mathbf{A}\in\{0,1\}^{n\times n}\), where \(\mathbf{A}_{ij}=1\) if and only if \((v_i,v_j)\in\mathcal{E}\).

\paragraph{Incomplete observations} 

In practice, both node attributes and edges may be partially missing, yielding incomplete observations \(\mathcal{G}_{\text{obs}}=(\mathcal{V},\mathcal{E}_{\text{obs}}, \mathbf{X}_{\text{obs}})\), where \(\mathcal{E}_{\text{obs}}\subseteq\mathcal{E}\) and \(\mathbf{X}_{\text{obs}}\) is the incomplete attribute matrix. The partially observed attributes and adjacency structure are described by  \(\mathbf{M}_X\in\{0,1\}^{n\times d}\) and \(\mathbf{M}_A\in\{0,1\}^{n\times n}\), respectively, as $\mathbf{X}_{\text{obs}}=\mathbf{M}_X\odot\mathbf{X},  \mathbf{A}_{\text{obs}}=\mathbf{M}_A\odot\mathbf{A}$, where \(\odot\) denotes the element-wise product. Unobserved entries in \(\mathbf{X}_{\text{obs}}\) and \(\mathbf{A}_{\text{obs}}\) (i.e., entries with mask values of \(0\)) are initially zero-filled and serve as placeholders during preprocessing.

\paragraph{Objective}  

Under the majority-normal assumption, the objective is to learn an anomaly-scoring function $f:(\mathbf{X}_{\text{obs}},\mathbf{A}_{\text{obs}},\mathbf{M}_X,\mathbf{M}_A) \rightarrow \mathbf{s}\in\mathbb{R}^{n}$, where a larger score \(s_i\) indicates a greater deviation of node \(v_i\) from the normal patterns embedded in the incomplete graph \(\mathcal{G}_{\text{obs}}\). During inference, the problem is formulated as a ranking task, with higher scores marking as anomalies~\cite{kaize2019dadan}.

\subsection{Dual-pathway Incomplete Graph Imputation}
\label{sec:dual-pathway}

Simultaneous incompleteness of node attributes and graph structure leads to mutual interference~\cite{huo2023t2gnn,fu2023towards}: attribute imputation relying on incomplete topology suffers from inaccurate neighborhood contexts, while structure completion based on incomplete or noisy attributes propagates reconstruction errors across the graph. To alleviate such interference, we propose a dual-pathway imputation module comprising: 1) a feature reconstruction pathway that imputes node attributes independently of structural information, and 2) a structure reconstruction pathway that enriches the observed sparse topology through global diffusion. The separately reconstructed attributes and structure are subsequently integrated to form a refined surrogate graph for improved downstream anomaly detection.

\subsubsection{Feature Reconstruction Pathway} 

To reconstruct missing node features, we employ a self-supervised MLP imputer. MLPs are robust to input sparsity and can flexibly model nonlinear relationships without relying on the graph topology~\cite{yoon2018gain}. Formally, given $\X_{\mathrm{obs}}$ and its binary mask $\M_X$, an MLP encoder $f_{\theta}$ is employed to reconstruct missing features as $\hat{\X}=f_{\theta}(\X_{\mathrm{obs}})$, where $f_{\theta}$ is trained by minimizing the mean squared error (MSE) between observed entries in the original attribute matrix $\X$ and their reconstructed $\hat{\X}$:
\begin{align}
    \mathcal{L}_{\text{feat}}
  = \operatorname{MSE}(\M_X \odot \hat{\X}, \M_X \odot \X).
\end{align}

\subsubsection{Structure Reconstruction Pathway} 

To recover the incomplete graph topology, we adopt deterministic Personalized PageRank (PPR) diffusion~\cite{park2019survey}. This diffusion propagates relationships beyond immediate neighbors on a global scale, thus alleviating node isolation caused by structural sparsity. Specifically, PPR iteratively diffuses structural information:
\begin{align}
\mathbf{P}^{(t+1)}=\beta\,\tilde{\A}\,\mathbf{P}^{(t)} + (1-\beta)\,\mathbf{I},\qquad
\tilde{\A} = \A_{\text{obs}} + \mathbf{I},
\end{align}
where $\beta$ is the teleport probability, and $t$ represents the iteration step. After convergence at step $T$, the final diffused structure is obtained as $\A_{\mathrm{ppr}}=\mathbf{P}^T$. The enhanced adjacency is then obtained by superimposing the diffusion result on the original observation:
\begin{align}
       \hat{\mathbf{A}} = \mathbf{A}_{\mathrm{obs}} + \mathbf{A}_{\mathrm{ppr}}.
\end{align}

The pair \((\hat{\mathbf{A}},\hat{\mathbf{X}})\) constitutes a densified surrogate graph used in the subsequent anomaly-scoring module.

\subsection{Joint Latent-Space Modeling}
\label{sec:joint-latent}

While the dual-pathway imputer reduces cross interference, it ignores the intrinsic co-dependencies between attributes and structure, i.e., a missing node feature can be inferred from the attributes of its neighbors, and conversely, a missing edge may be recovered using similarity in node features. To capture these dependencies, we embed both reconstructed views into a shared latent space, allowing each view to mutually refine and complement the other. Then we impose 1) a spherical prior to compact normal nodes and 2) a reconstruction objective to preserve fine-grained semantics.

\subsubsection{Latent-Space Fusion and Spherical Constraint} 

Formally, we denote the GCN projector by \(g_{\gamma}\), an \(L\)-layer GCN with propagation rules:
\[
\mathbf{Z}^{(0)} = \hat{\X},\quad
\mathbf{Z}^{(l)} = \sigma\bigl(\hat{\A}\,\mathbf{Z}^{(l-1)}\,W^{(l)}\bigr)\quad(l=1,\dots,L),
\]
where \(W^{(l)}\in\mathbb{R}^{d_{l-1}\times d_l}\) and \(\sigma(\cdot)\) is ReLU. The fused node embedding is then simply
$
\mathbf{Z} = g_{\gamma}(\hat{\X},\,\hat{\A}) = \mathbf{Z}^{(L)}
$.

To shape the empirical distribution of $\mathbf{Z}$, a Sinkhorn divergence $\mathcal{L}_{\mathrm{dist}}$ is minimised between $\mathbf{Z}$ and a truncated Gaussian prior \(\mathcal{N}_r(\mathbf{0},\mathbf{I})\) supported on the latent ball ${\| z\|_2 \leq r}$:
\begin{align}
\mathcal{L}_{\mathrm{dist}} = \operatorname{Sinkhorn}(\mathbf{Z}, \mathcal{N}_{r}(\mathbf{0}, \mathbf{I})).
\end{align}
Under the assumption that most nodes are normal, this constraint compacts their embeddings within radius \(r\) and implicitly reserves the exterior for anomalies. Unlike point-to-center objectives (e.g., SVDD~\cite{ruff2018deep}), Sinkhorn offers global distribution matching: it aligns the entire embedding cloud to the truncated-Gaussian prior, encouraging a meaningful spread of normal patterns inside the latent ball and preventing trivial collapse.

\subsubsection{Latent Semantics Preservation via Reconstruction}

To tether the compact latent space to the input’s fine-grained semantics, we append a two-layer MLP recovery decoder \(r_{\omega}\) seeking to reconstruct node attributes from the embeddings, i.e., $\tilde{\X} = r_{\omega}(\mathbf{Z})$, trained under the MSE objective:
\begin{align}
    \mathcal{L}_{\mathrm{recon}} = \operatorname{MSE}(\M_X \odot \tilde{\X}, \M_X \odot \X).
\end{align}
Jointly minimising \(\mathcal{L}_{\mathrm{recon}}\) alongside the spherical Sinkhorn loss preserves attribute–structure details in each embedding for faithful reconstruction and prevents collapse.

\subsection{Graph Pseudo-Anomaly Generation}
\label{sec:pseudo-anomaly}

The scarcity of true anomalies and the absence of labels can cause imputation bias: a model trained almost exclusively on majority-normal data tends to reconstruct anomalous nodes as if they were normal, thereby erasing distinctive signals and degrading detection performance. To counteract this effect we aim to generate sufficient pseudo‑anomalies. The idea is to sample latent codes that lie just outside the compact normal region, decode them into node attributes via the decoder \(r_{\omega}\), endow them with a lightweight internal topology, and append the resulting subgraph to the training graph as hard negative examples.

\subsubsection{Latent-Space Sampling}

After the latent‑space training in Section~\ref{sec:joint-latent}, normal embeddings are concentrated inside a ball of radius \(r\). Rather than sampling latent codes arbitrarily far from this region, which can produce trivial or unrealistic outliers, we focus on the margin where boundary refinement is most critical. Specifically, we sample \(M\) latent vectors \(\{\,\mathbf{z}^{(a)}_i\}_{i=1}^{M}\) uniformly from an annular shell:
\begin{align}
   \mathcal{S}_{\text{shell}}
  = \bigl\{\mathbf{z}\in\mathbb{R}^{d_z}\mid r_a<\|\mathbf{z}\|_2<r_b\bigr\}. 
\end{align}
Vectors in this shell remain close enough to the normal manifold to be decoded into realistic feature–structure patterns, yet lie just outside the compact region, providing hard negative examples that sharpen the model’s decision boundary. To avoid excessive hyper-parameter tuning, we set $r_a = 1.2\,r$ and $r_b = 2r$ for all experiments, which we find to work robustly across datasets.

\subsubsection{Synthetic Graph Construction}

Each sampled vector is then mapped to the original attribute space through the decoder, i.e., $\tilde{x}^{(a)} = r_{\omega}(z_i^{(a)})$. Since the decoder has been trained to reconstruct authentic node features, the resulting $\tilde{x}_i^{(a)}$ resembles plausible but atypical nodes.

However, a key challenge is how to embed the pseudo-anomaly nodes into the training graph, without introducing spurious connections that would either corrupt the normal manifold, or yield unrealistic, uninformative structures (e.g., if left isolated). To address this, we create an internal pseudo-anomaly subgraph whose edges reflect feature similarity, while keeping it disconnected from the real nodes.

Specifically, we compute pairwise cosine similarity among the decoded features and connect node pair $(i, j)$ whenever $\mathrm{cos}(\tilde{x}_i^{(a)}, \tilde{x}_j^{(a)}) \geq \tau_a$, yielding an adjacency matrix $\mathbf{S}^{(a)} \in {(0, 1)}^{M\times M}$. To align with the characteristics of $\mathbf{A}_{\text{obs}}$, we perform row-wise min-max normalization on $\mathbf{S}^{(a)}$, followed by binarization with a fixed threshold $\tau_a=0.5$:
\begin{align}
    \mathbf{A}^{(a)}_{ij}
  = \mathbb{I}\!\Bigl(
      \frac{\mathbf{S}^{(a)}_{ij}
            -\min\nolimits_{k}\mathbf{S}^{(a)}_{ik}}
           {\max\nolimits_{k}\mathbf{S}^{(a)}_{ik}
            -\min\nolimits_{k}\mathbf{S}^{(a)}_{ik}}
      \ge \tau_a
    \Bigr).
\end{align}
This ensures all node-specific similarities are mapped to $[0,1]$, preserving local topological rankings while enabling direct comparison with $\mathbf{A}_{\text{obs}}$. 

\begin{table*}[t]
    \centering
    \resizebox{\linewidth}{!}{
    \begin{tabular}{c | c | c c c c c c c} 
    
    \hline
    DI Method & GAD Method  & Cora & Citeseer & Books & Disney & Flickr &  ACM  & Reddit   \\

    \hline
    \multirow{5}{*}{Mean} & COLA (TNNLS 2022) \cite{liu2022cola} & 0.46$\pm$0.02 & 0.44$\pm$0.02 & 0.50$\pm$0.04 & 0.48$\pm$0.09 & 0.50$\pm$0.01 & 0.45$\pm$0.01 & 0.51$\pm$0.02 \\
    
        & PREM (ICDM 2023) \cite{pan2023prem} & 0.63$\pm$0.02 & 0.67$\pm$0.01 & 0.36$\pm$0.02 & 0.29$\pm$0.07 & 0.58$\pm$0.01 & 0.59$\pm$0.01 & 0.54$\pm$0.01 \\
        
        & GRADATE (AAAI 2023) \cite{duan2023gradate} & 0.49$\pm$0.03 & 0.48$\pm$0.02 & 0.52$\pm$0.08 & 0.49$\pm$0.09 & 0.46$\pm$0.02 & 0.46$\pm$0.05 & 0.51$\pm$0.05 \\
        
        & ADA-GAD (AAAI 2024) \cite{he2024ada} & 0.84$\pm$0.01 & 0.90$\pm$0.01 & 0.41$\pm$0.05 & 0.40$\pm$0.04 & 0.70$\pm$0.01 & 0.78$\pm$0.00 & 0.56$\pm$0.01 \\
        
        & DiffGAD (ICLR 2025) \cite{li202diffgad} & 0.52$\pm$0.05 & 0.50$\pm$0.08 & 0.51$\pm$0.07 & 0.49$\pm$0.06 & 0.51$\pm$0.01 & 0.60$\pm$0.01 & 0.52$\pm$0.05 \\
    \hline
    \multirow{5}{*}{MissForest} & COLA (TNNLS 2022) \cite{liu2022cola} & 0.52$\pm$0.01 & 0.49$\pm$0.02 & 0.49$\pm$0.10 & 0.55$\pm$0.12 & 0.64$\pm$0.01 & 0.48$\pm$0.02 & 0.52$\pm$0.01 \\
    
        & PREM (ICDM 2023) \cite{pan2023prem} & 0.65$\pm$0.01 & 0.65$\pm$0.01 & 0.40$\pm$0.04 & 0.37$\pm$0.08 & 0.75$\pm$0.01 & 0.66$\pm$0.01 & 0.55$\pm$0.00 \\
        
        & GRADATE (AAAI 2023) \cite{duan2023gradate} & 0.52$\pm$0.03 & 0.49$\pm$0.01 & 0.54$\pm$0.10 & 0.55$\pm$0.05 & 0.64$\pm$0.01 & 0.48$\pm$0.07 & 0.55$\pm$0.03 \\
        
        & ADA-GAD (AAAI 2024) \cite{he2024ada} & 0.84$\pm$0.01 & 0.70$\pm$0.13 & 0.54$\pm$0.12 & 0.45$\pm$0.06 & 0.69$\pm$0.01 & 0.78$\pm$0.01 & 0.49$\pm$0.05 \\
        
        & DiffGAD (ICLR 2025) \cite{li202diffgad} & 0.50$\pm$0.10 & 0.55$\pm$0.03 & 0.48$\pm$0.05 & 0.45$\pm$0.10 & 0.62$\pm$0.01 & 0.48$\pm$0.01 & 0.48$\pm$0.04 \\
    \hline
    \multirow{5}{*}{GAE} & COLA (TNNLS 2022) \cite{liu2022cola} & 0.53$\pm$0.04 & 0.53$\pm$0.01 & 0.52$\pm$0.06 & 0.58$\pm$0.11 & 0.48$\pm$0.03 & 0.51$\pm$0.01 & 0.50$\pm$0.02 \\
    
        & PREM (ICDM 2023) \cite{pan2023prem} & 0.41$\pm$0.08 & 0.32$\pm$0.02 & 0.40$\pm$0.04 & 0.46$\pm$0.13 & 0.55$\pm$0.02 & 0.48$\pm$0.06 & 0.51$\pm$0.01 \\
        
        & GRADATE (AAAI 2023) \cite{duan2023gradate} & 0.49$\pm$0.02 & 0.50$\pm$0.02& 0.58$\pm$0.05 & 0.55$\pm$0.17  & 0.49$\pm$0.01 & 0.50$\pm$0.02 & 0.49$\pm$0.03 \\
        
        & ADA-GAD (AAAI 2024) \cite{he2024ada} & 0.66$\pm$0.02 & 0.64$\pm$0.04 & 0.57$\pm$0.09 & 0.51$\pm$0.11 & 0.56$\pm$0.01 & 0.64$\pm$0.01 & 0.55$\pm$0.01 \\
        
        & DiffGAD (ICLR 2025) \cite{li202diffgad} & 0.66$\pm$0.02 & 0.61$\pm$0.01  & 0.49$\pm$0.03 & 0.46$\pm$0.08 & 0.61$\pm$0.01 & 0.57$\pm$0.01 & 0.51$\pm$0.06 \\
    \hline

    \multirow{5}{*}{ASD-VAE} & COLA (TNNLS 2022) \cite{liu2022cola} & 0.54$\pm$0.02 & 0.49$\pm$0.03 & 0.66$\pm$0.01 & 0.54$\pm$0.01 & 0.37$\pm$0.11 & 0.53$\pm$0.05 & 0.50$\pm$0.02 \\
    
    &  PREM (ICDM 2023) \cite{pan2023prem} & 0.64$\pm$0.00 & 0.66$\pm$0.02 & 0.58$\pm$0.04 & 0.59$\pm$0.01 & 0.35$\pm$0.05 & 0.40$\pm$0.03 & 0.56$\pm$0.01 \\
    
    &  GRADATE (AAAI 2023) \cite{duan2023gradate} & 0.51$\pm$0.03 & 0.49$\pm$0.03 & 0.58$\pm$0.03 & 0.51$\pm$0.05 & 0.39$\pm$0.22 & 0.49$\pm$0.06 & 0.49$\pm$0.01 \\
    
    &  ADA-GAD (AAAI 2024) \cite{he2024ada} & 0.83$\pm$0.01 & 0.90$\pm$0.01 & \textbf{0.70$\pm$0.00} & 0.78$\pm$0.01 & 0.41$\pm$0.09 & 0.43$\pm$0.06 & 0.56$\pm$0.00 \\
    
    &  DiffGAD (ICLR 2025) \cite{li202diffgad} & 0.79$\pm$0.01 & 0.50$\pm$0.00 & 0.50$\pm$0.00 & 0.50$\pm$0.00 & 0.48$\pm$0.08 & 0.50$\pm$0.00 & 0.55$\pm$0.00 \\ \cline{2-8}
    \hline
    
    \multirow{5}{*}{GAIN} & COLA (TNNLS 2022) \cite{liu2022cola} & 0.53$\pm$0.01 & 0.50$\pm$0.03 & 0.50$\pm$0.08 & 0.58$\pm$0.10 & 0.67$\pm$0.01 & 0.56$\pm$0.01 & 0.51$\pm$0.02 \\
    
        & PREM (ICDM 2023) \cite{pan2023prem} & 0.62$\pm$0.02 & 0.66$\pm$0.01 & 0.42$\pm$0.02 & 0.61$\pm$0.06 & 0.56$\pm$0.04 & 0.57$\pm$0.01 & 0.55$\pm$0.01 \\
        
        & GRADATE (AAAI 2023) \cite{duan2023gradate} & 0.49$\pm$0.03 & 0.51$\pm$0.05 & 0.58$\pm$0.08 & 0.57$\pm$0.07 & 0.61$\pm$0.03 & 0.52$\pm$0.03 & 0.53$\pm$0.04 \\
        
        & ADA-GAD (AAAI 2024) \cite{he2024ada} & 0.84$\pm$0.01 & 0.90$\pm$0.01 & 0.60$\pm$0.04 & 0.44$\pm$0.12 & 0.74$\pm$0.01 & 0.82$\pm$0.01 & 0.47$\pm$0.01 \\
        
        & DiffGAD (ICLR 2025) \cite{li202diffgad} & 0.47$\pm$0.08 & 0.50$\pm$0.01 & 0.51$\pm$0.01 & 0.46$\pm$0.07 & 0.65$\pm$0.01 & 0.53$\pm$0.01 & 0.49$\pm$0.04 \\
    \hline
    
    \multicolumn{2}{c|}{M$^2$V-UGAD (ours)} & \textbf{0.93$\pm$0.02} & \textbf{0.92$\pm$0.02} & 0.63$\pm$0.02 & \textbf{0.81$\pm$0.05} & \textbf{0.93$\pm$0.01} & \textbf{0.92$\pm$0.01} & \textbf{0.58$\pm$0.02} \\

    \hline
    \end{tabular}}
    \caption{Detection performance comparison with the existing GAD methods under 30$\%$ missing rate in terms of AUROC.}
    \label{table:comparison_study}
\end{table*}

Finally, we build a block‑diagonal augmented graph:
\begin{align}
\mathbf{A}_{\text{aug}}
  =\begin{bmatrix}
      \mathbf{A}_{\text{obs}} & \mathbf{0}\\
      \mathbf{0} & \mathbf{A}^{(a)}
    \end{bmatrix},
\qquad
\mathbf{X}_{\text{aug}}
  =\begin{bmatrix}
      \mathbf{X}_{\text{obs}}\\
      \tilde{\mathbf{X}}^{(a)}
    \end{bmatrix}.
\end{align}
The synthetic subgraph thus supplies realistic, self‑contained pseudo‑anomalies that act as explicit contrastive signals, sharpening the decision boundary of the downstream anomaly detector without introducing spurious mixed‑type edges.

\subsection{Training Pipeline and Anomaly Scoring}
\label{sec:training}

Overall, our training proceeds in two stages. In the pre-training stage, we jointly optimize $\LL_{\mathrm{pretrain}}$ on the original graph $(\X_{\mathrm{obs}}, \A_{\mathrm{obs}})$:
\begin{align}
\LL_{\mathrm{pretrain}} = \LL_{\mathrm{dist}} +\alpha \LL_{\mathrm{feat}} +   \lambda \LL_{\mathrm{recon}},
\end{align}
where $\alpha, \lambda$ are hyperparameters balancing their contributions. In the fine-tuning stage, we augment the data with $M=\eta n$ pseudo-anomalies ($\eta$ denotes the fraction of pseudo-anomaly nodes among all training samples), to obtain ( $\X_{\mathrm{aug}}$, $\A_{\mathrm{aug}}$) and re-optimise:
\begin{align}
    \LL_{\mathrm{finetune}} = \LL_{\mathrm{dist}}^{'} +\alpha \LL_{\mathrm{feat}} +   \lambda \LL_{\mathrm{recon}},
\end{align}
where
\begin{align}
\LL_{\mathrm{dist}}' 
&= \mathrm{Sinkhorn}\bigl(\mathbf{Z}_{\mathrm{pseudo}},\,\mathcal{N}_{\{r_a<\|z\|_2\le r_b\}}(0,\mathbf{I})\bigr) \notag \\
&+ \mathrm{Sinkhorn}\bigl(\mathbf{Z},\,\mathcal{N}_r(0,\mathbf{I})\bigr)
\end{align}
compacts normal embeddings inside the radius $r$ ball while pushing pseudo-anomaly embeddings toward the outer shell \(\{r_a<\|z\|_2\le r_b\}\).
During inference, each node $v_i$ is processed through the trained imputers and projector to produce its latent embedding $z_i$. Its anomaly score is defined as $s_i = \|z_i\|_2$, where larger values indicate greater deviation from the learned normal region.

%% file: experiment_v1.tex
\subsection{Datasets}

We perform experiments on seven real-world attributed graphs spanning diverse application domains. Four of these datasets (Cora, Citeseer, Flickr, and ACM) do not include ground-truth anomaly labels and are hence used for anomaly injection, while the other three (Books, Disney, and Reddit) contain inherent anomalies derived from user behavior or content semantics. For the injection datasets, we follow the CoLA protocol~\cite{liu2022cola} to synthesize structural anomalies by randomly rewiring edges among a subset of nodes and contextual anomalies by replacing selected nodes’ feature vectors with those of semantically distant peers. To simulate real-world missingness, we randomly mask 30\% of nodes and 30\% of edges in every graph for our main experiments; Table~\ref{table:comparison_missing_rate} explores additional masking rates to evaluate the robustness of our method under varying degrees of incompleteness. The summary statistics of all datasets are presented in the Supplementary Material.

\begin{table*}[t]
    \centering
    \resizebox{0.9\linewidth}{!}{
    \begin{tabular}{c | c | c | c c c c c } 
    
    \hline
    Dataset & DI Method & GAD Method  & $10\%$ & $20\%$ & $30\%$  & $40\%$ & $50\%$    \\

    \hline
    \multirow{26}{*}{Cora}
     & \multirow{5}{*}{Mean} & COLA (TNNLS 2022) \cite{liu2022cola} & 0.53$\pm$0.02 & 0.49$\pm$0.02 & 0.46$\pm$0.02 & 0.44$\pm$0.01 & 0.47$\pm$0.02 \\
     
        &  &  PREM (ICDM 2023) \cite{pan2023prem} & 0.70$\pm$0.01 & 0.66$\pm$0.02 & 0.63$\pm$0.02 & 0.62$\pm$0.02 & 0.65$\pm$0.01 \\
        
        &  &  GRADATE (AAAI 2023) \cite{duan2023gradate} & 0.50$\pm$0.04 & 0.52$\pm$0.02 & 0.49$\pm$0.03 & 0.49$\pm$0.02 & 0.49$\pm$0.02 \\
        
        &  &  ADA-GAD (AAAI 2024) \cite{he2024ada} & 0.85$\pm$0.01 & 0.85$\pm$0.00 & 0.84$\pm$0.01 & 0.82$\pm$0.01 & 0.82$\pm$0.01 \\
        
        &  &  DiffGAD (ICLR 2025) \cite{li202diffgad} & 0.53$\pm$0.06 & 0.47$\pm$0.01 & 0.52$\pm$0.05 & 0.53$\pm$0.08 & 0.46$\pm$0.08 \\ \cline{2-8}
        
     & \multirow{5}{*}{MissForest} & COLA (TNNLS 2022) \cite{liu2022cola} & 0.53$\pm$0.01 & 0.56$\pm$0.01 & 0.52$\pm$0.01 & 0.53$\pm$0.01 & 0.55$\pm$0.01 \\
     
        &  &  PREM (ICDM 2023) \cite{pan2023prem} & 0.71$\pm$0.01 & 0.70$\pm$0.01 & 0.65$\pm$0.01 & 0.64$\pm$0.00 & 0.64$\pm$0.01 \\
        
        &  &  GRADATE (AAAI 2023) \cite{duan2023gradate} & 0.52$\pm$0.02 & 0.48$\pm$0.01 & 0.52$\pm$0.03 & 0.52$\pm$0.03 & 0.53$\pm$0.02 \\
        
        &  &  ADA-GAD (AAAI 2024) \cite{he2024ada} & 0.86$\pm$0.01 & 0.85$\pm$0.01 & 0.84$\pm$0.01 & 0.80$\pm$0.00 & 0.77$\pm$0.01 \\
        
        &  &  DiffGAD (ICLR 2025) \cite{li202diffgad} & 0.45$\pm$0.07 & 0.51$\pm$0.07 & 0.50$\pm$0.10 & 0.53$\pm$0.10 & 0.49$\pm$0.09 \\ \cline{2-8}
        
     & \multirow{5}{*}{GAE} & COLA (TNNLS 2022) \cite{liu2022cola} & 0.52$\pm$0.02 & 0.53$\pm$0.04 & 0.53$\pm$0.04 & 0.56$\pm$0.01 & 0.51$\pm$0.02 \\
     
        &  &  PREM (ICDM 2023) \cite{pan2023prem} & 0.39$\pm$0.06 & 0.32$\pm$0.03 & 0.41$\pm$0.08 & 0.40$\pm$0.06 & 0.35$\pm$0.06 \\
        
        &  &  GRADATE (AAAI 2023) \cite{duan2023gradate} & 0.52$\pm$0.02 & 0.51$\pm$0.02 & 0.49$\pm$0.02 & 0.52$\pm$0.04 & 0.47$\pm$0.03 \\
        
        &  &  ADA-GAD (AAAI 2024) \cite{he2024ada} & 0.67$\pm$0.01 & 0.65$\pm$0.01 & 0.66$\pm$0.02 & 0.60$\pm$0.02 & 0.58$\pm$0.01 \\
        
        &  &  DiffGAD (ICLR 2025) \cite{li202diffgad} & 0.68$\pm$0.02 & 0.66$\pm$0.02 & 0.66$\pm$0.02 & 0.59$\pm$0.01 & 0.64$\pm$0.01 \\ \cline{2-8}

    & \multirow{5}{*}{ASD-VAE} & COLA (TNNLS 2022) \cite{liu2022cola} & 0.51$\pm$0.02 & 0.52$\pm$0.02 & 0.54$\pm$0.02 & 0.52$\pm$0.00 & 0.55$\pm$0.01 \\
    
        &  &  PREM (ICDM 2023) \cite{pan2023prem} & 0.69$\pm$0.01 & 0.66$\pm$0.01 & 0.64$\pm$0.00 & 0.63$\pm$0.03 & 0.60$\pm$0.02 \\
        
        &  &  GRADATE (AAAI 2023) \cite{duan2023gradate} & 0.50$\pm$0.03 & 0.51$\pm$0.01 & 0.51$\pm$0.03 & 0.52$\pm$0.03 & 0.50$\pm$0.02 \\
        
        &  &  ADA-GAD (AAAI 2024) \cite{he2024ada} & 0.86$\pm$0.01 & 0.84$\pm$0.01 & 0.83$\pm$0.01 & 0.83$\pm$0.01 & 0.82$\pm$0.01 \\
        
        &  &  DiffGAD (ICLR 2025) \cite{li202diffgad} & 0.78$\pm$0.02 & 0.79$\pm$0.01 & 0.79$\pm$0.01 & 0.50$\pm$0.01 & 0.50$\pm$0.01 \\ \cline{2-8}
        
     & \multirow{5}{*}{GAIN} & COLA (TNNLS 2022) \cite{liu2022cola} & 0.54$\pm$0.02 & 0.53$\pm$0.01 & 0.53$\pm$0.01 & 0.55$\pm$0.02 & 0.57$\pm$0.04 \\
     
        &  &  PREM (ICDM 2023) \cite{pan2023prem} & 0.69$\pm$0.01 & 0.64$\pm$0.01 & 0.62$\pm$0.02 & 0.58$\pm$0.02 & 0.63$\pm$0.01 \\
        
        &  &  GRADATE (AAAI 2023) \cite{duan2023gradate} & 0.51$\pm$0.03 & 0.50$\pm$0.03 & 0.49$\pm$0.03 & 0.50$\pm$0.01 & 0.53$\pm$0.02 \\
        
        &  &  ADA-GAD (AAAI 2024) \cite{he2024ada} & 0.86$\pm$0.01 & 0.84$\pm$0.01 & 0.84$\pm$0.01 & 0.83$\pm$0.00 & 0.83$\pm$0.01 \\
        
        &  &  DiffGAD (ICLR 2025) \cite{li202diffgad} & 0.55$\pm$0.07 & 0.60$\pm$0.09 & 0.47$\pm$0.08 & 0.51$\pm$0.06 & 0.50$\pm$0.01 \\ \cline{2-8}

        &  \multicolumn{2}{c|}{M$^2$V-UGAD (ours)} & \textbf{0.93$\pm$0.01} & \textbf{0.93$\pm$0.02}  & \textbf{0.93$\pm$0.02} & \textbf{0.92$\pm$0.01} & \textbf{0.92$\pm$0.01} \\
        
    \hline
    \end{tabular}}
        \caption{Detection performance comparison with the existing GAD methods under various missing rates in terms of AUROC.}
    \label{table:comparison_missing_rate}
\end{table*}

\begin{figure*}[htbp]
    \includegraphics[width=1\linewidth]{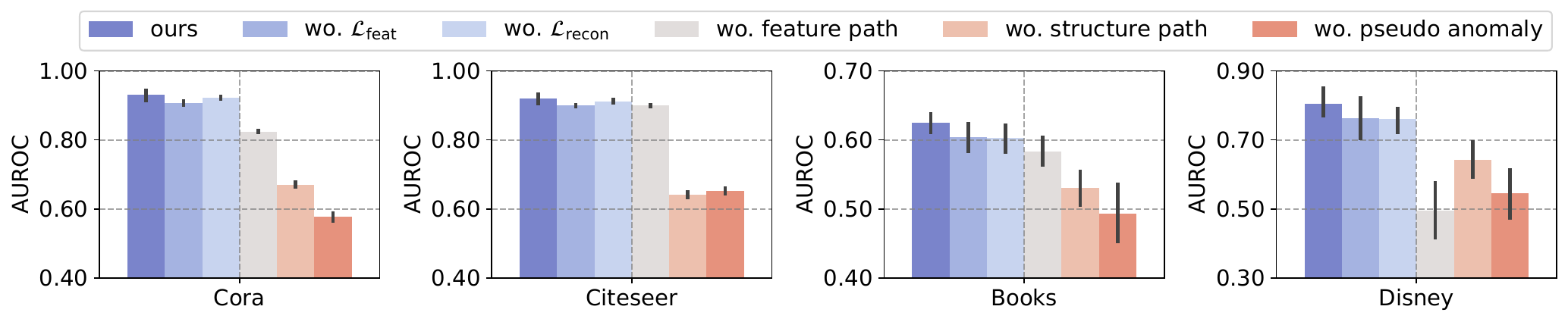}
    \caption{Ablation experiments of M$^2$V-UGAD under 30\% missing rate in terms of AUROC.}
    \label{fig:ablation}
\end{figure*}

\subsection{Baseline Methods and Implementation Details}

We compare M$^2$V-UGAD against five recent unsupervised GAD models, including CoLA~\cite{liu2022cola}, PREM~\cite{pan2023prem}, GRADATE~\cite{duan2023gradate}, ADA-GAD~\cite{he2024ada}, and DiffGAD~\cite{li202diffgad}. None of these detectors natively supports graphs with missing nodes or edges; therefore, we adopt a two-stage ``impute-then-detect'' pipeline. In the first stage, we restore missing values using one of five data imputation (DI) methods (Mean Filling, MissForest~\cite{stekhoven2011missforest}, GAIN~\cite{yoon2018gain}, GAE~\cite{kipf2016variational}, or ASD-VAE~\cite{jiang2024incomplete}). In the second stage, we apply each detector to each imputed graph. All methods are evaluated using the Area Under the ROC Curve (AUROC) with performance averaged over 5 independent runs. Detailed descriptions of each detector, imputer, pairing protocol, and the full M$^2$V-UGAD configuration are provided in the Supplementary Material.  

\subsection{Comparison Study}

Table~\ref{table:comparison_study} summarizes the AUROC scores for all methods across seven datasets at a 30\% missingness, while Table~\ref{table:comparison_missing_rate} further reports performance on the Cora dataset with varying masking ratios. Overall, M$^2$V-UGAD consistently outperforms all baselines across diverse experimental setups and the majority of datasets.

The suboptimal performance of the baselines stems from the limitations of the imputation methods. Mean Filling, MissForest, and GAIN treat nodes independently and neglect the structural relationships among nodes, resulting in large imputation errors under severe incompleteness. While GAE and ASD-VAE leverage graph topology, they implicitly assume that the observed structural information is complete and reliable. When edges are also missing, structural noise adversely affects attribute imputation due to mutual interference between structure and node attributes. Incomplete edges can further disrupt standard message-passing mechanisms. Even for low imputation error, the predominance of normal nodes introduces imputation bias, diluting the distinctive signals required for accurate anomaly detection.

\begin{figure*}[t]
\begin{subfigure}[t]{1\linewidth}
    \centering
    \includegraphics[height=1cm]{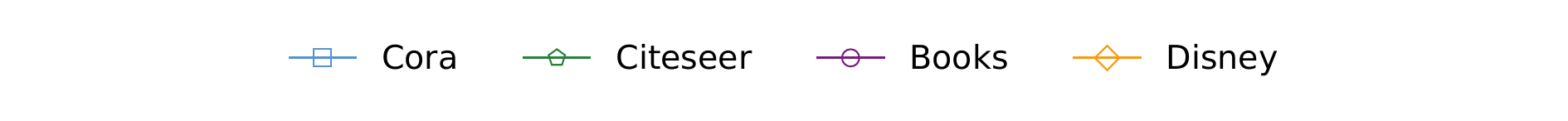}
\end{subfigure}
\vspace{-0.7cm}
\newline
\begin{subfigure}[b]{.245\linewidth}
    \includegraphics[height=2.85cm]{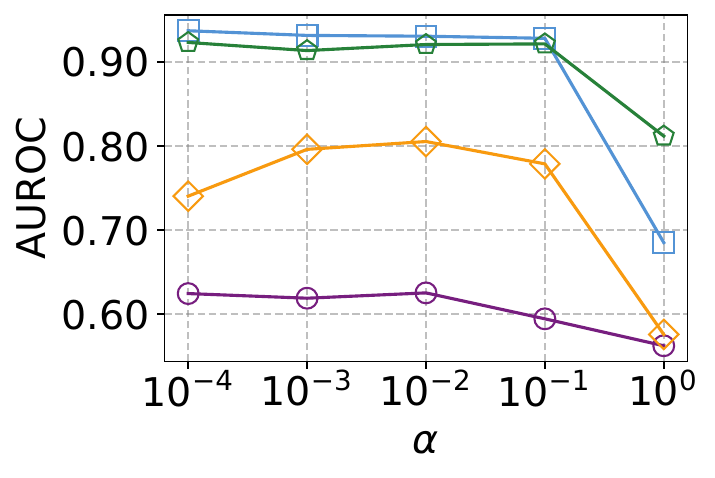} 
    \centering
    \caption{Imputation loss weight $\alpha$}
    \label{fig:sen1}
\end{subfigure}
\begin{subfigure}[b]{.245\linewidth}
    \includegraphics[height=2.85cm]{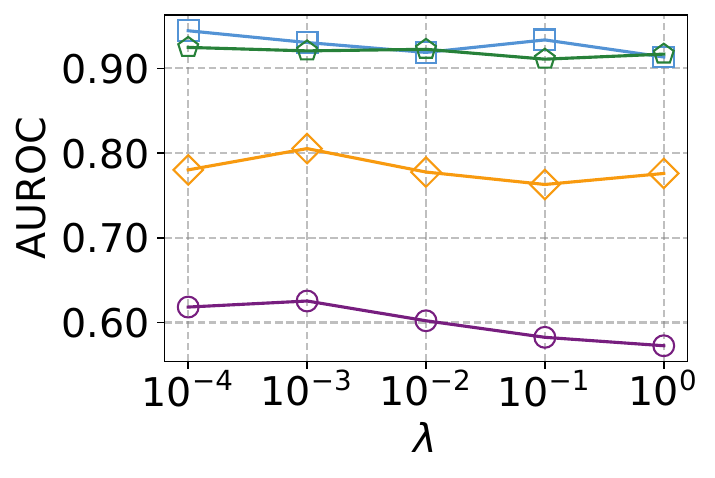} 
    \centering
    \caption{Reconstruction loss weight $\lambda$}
    \label{fig:sen2}
\end{subfigure}
\begin{subfigure}[b]{.245\linewidth}
    \includegraphics[height=2.85cm]{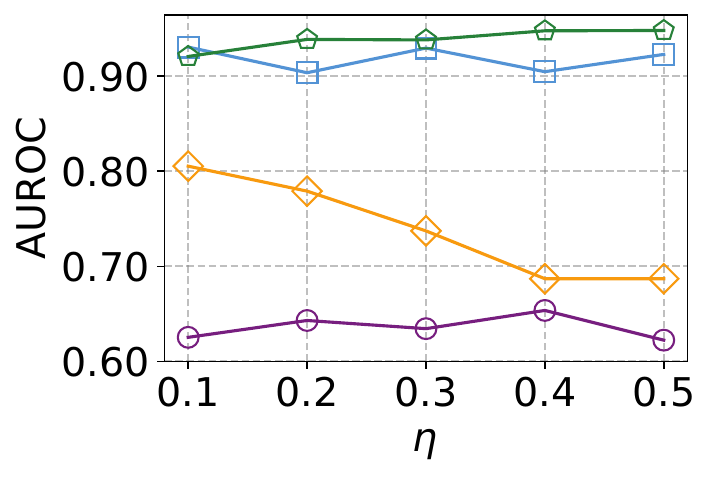} 
    \centering
    \caption{Pseudo-anomaly ratio $\eta$}
    \label{fig:sen3}
\end{subfigure}
\begin{subfigure}[b]{.245\linewidth}
    \includegraphics[height=2.85cm]{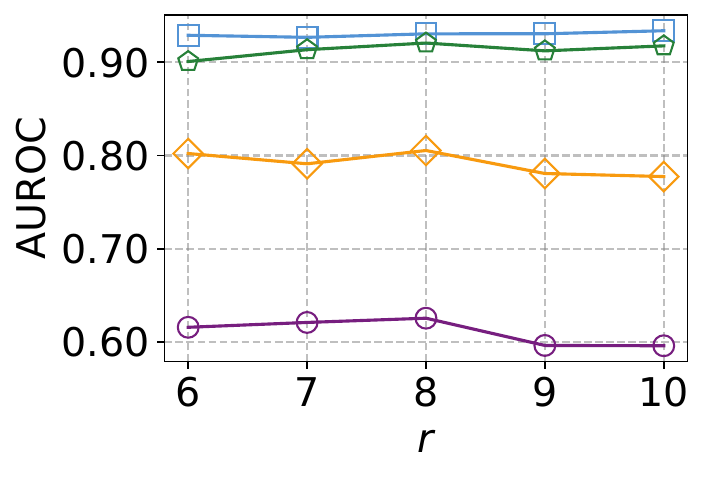} 
    \centering
    \caption{Latent-sphere radius $r$}
    \label{fig:sen4}
\end{subfigure}
\caption{Sensitivity analysis of M$^2$V-UGAD under 30$\%$ missing rate in terms of AUROC.}
\label{fig:hyper_param}
\end{figure*}

\begin{figure*}[htbp]
\hfill
\begin{subfigure}[b]{.16\linewidth}
    \includegraphics[height=1.8cm]{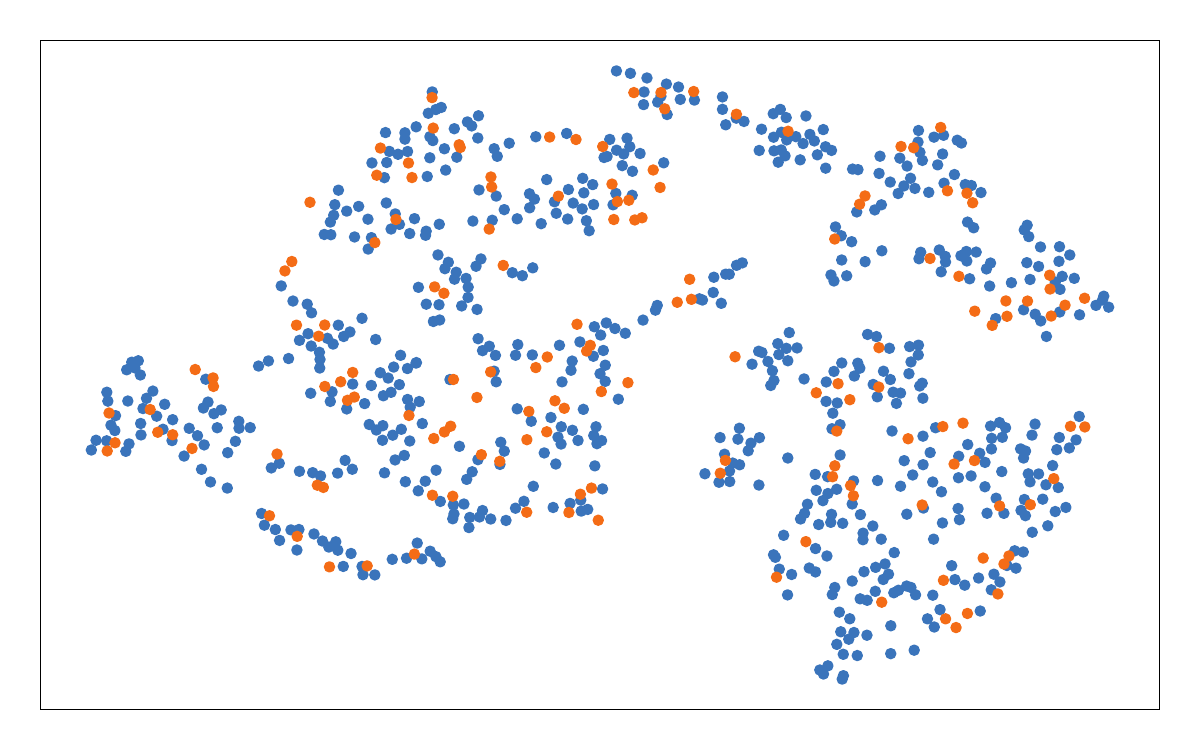} 
    \centering
    \caption{COLA}
    \label{fig:cola_visualization}
\end{subfigure}
\begin{subfigure}[b]{.16\linewidth}
    \includegraphics[height=1.8cm]{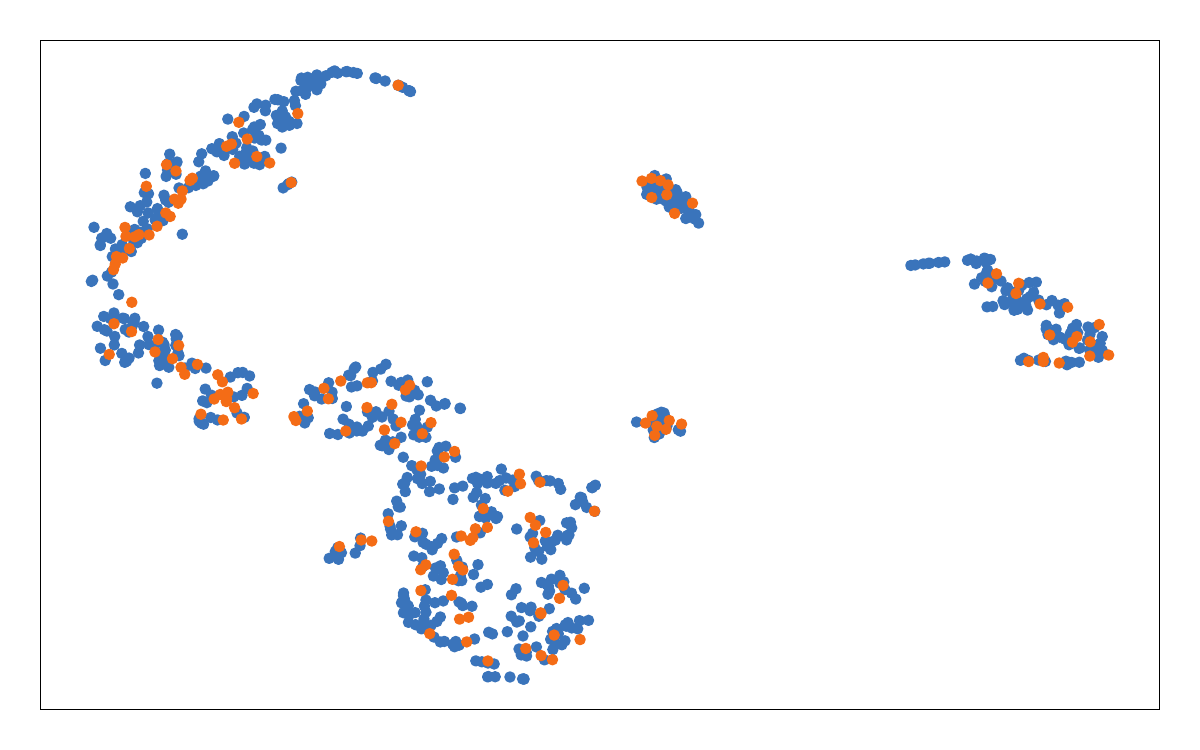} 
    \centering
    \caption{PREM}
    \label{fig:prem_visualization}
\end{subfigure}
\begin{subfigure}[b]{.16\linewidth}
    \includegraphics[height=1.8cm]{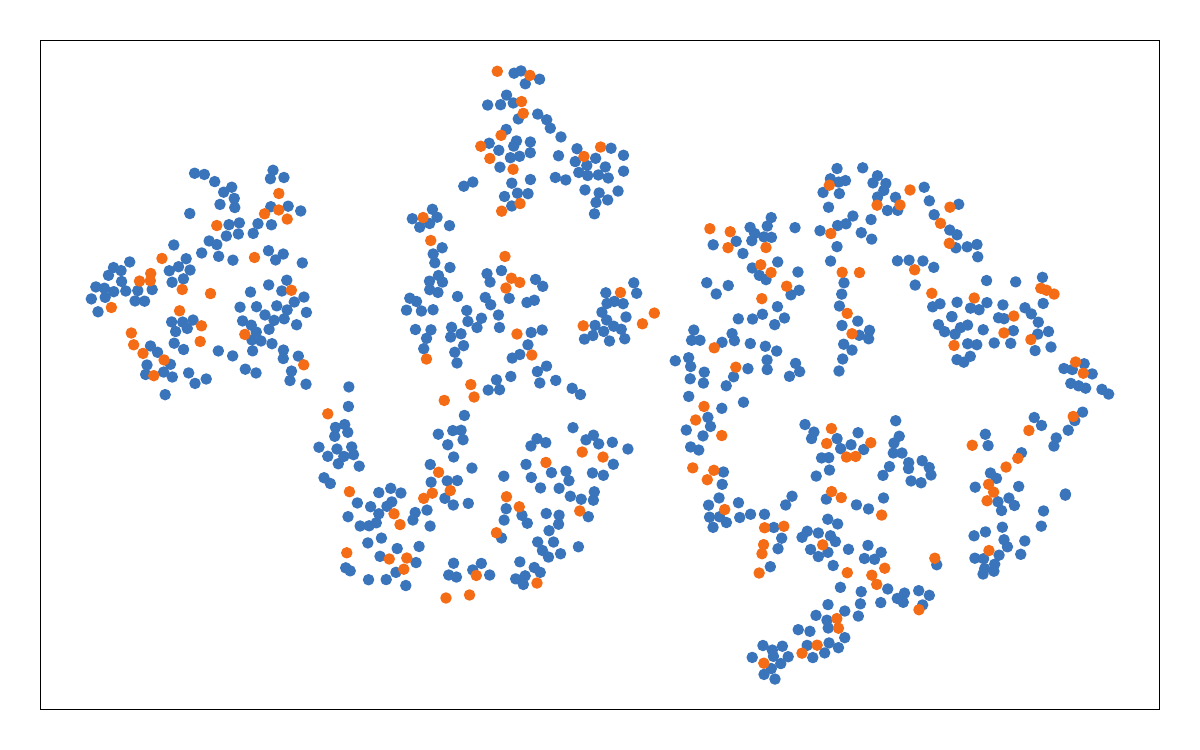} 
    \centering
    \caption{GRADATE}
    \label{fig:gradate_visualization}
\end{subfigure}
\begin{subfigure}[b]{.16\linewidth}
    \includegraphics[height=1.8cm]{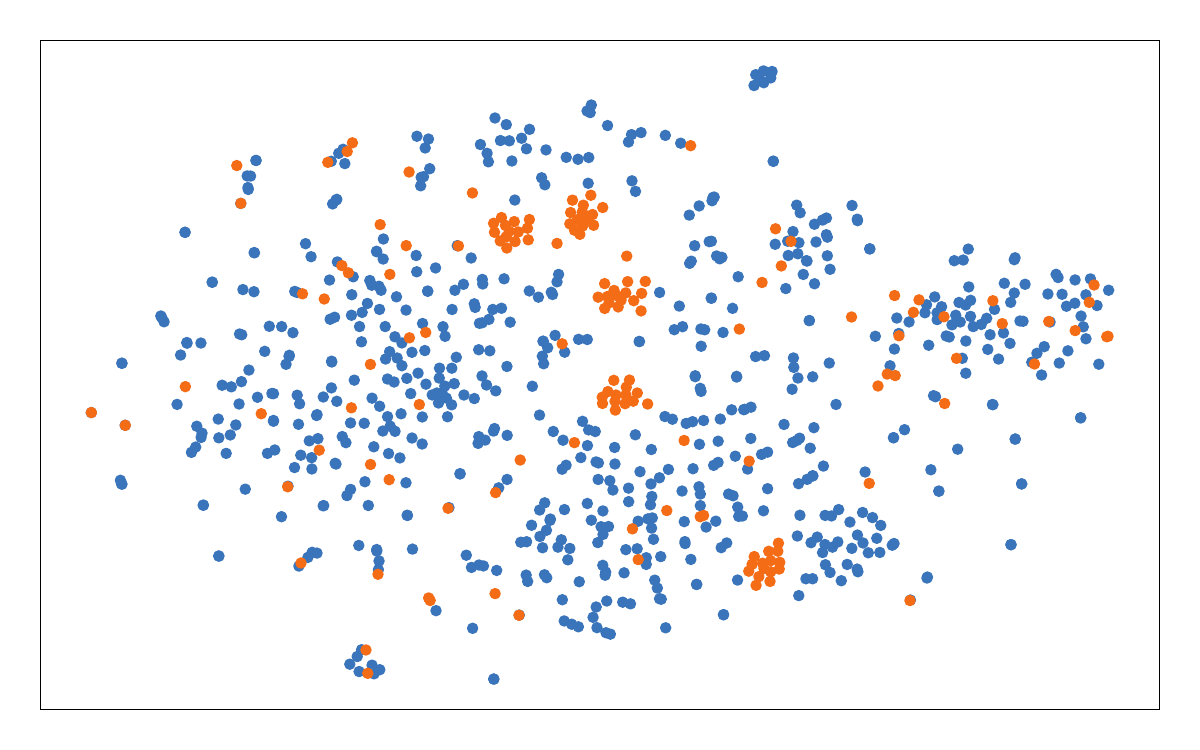} 
    \centering
    \caption{ADA-GAD}
    \label{fig:ada_visualization}
\end{subfigure}
\begin{subfigure}[b]{.16\linewidth}
    \includegraphics[height=1.8cm]{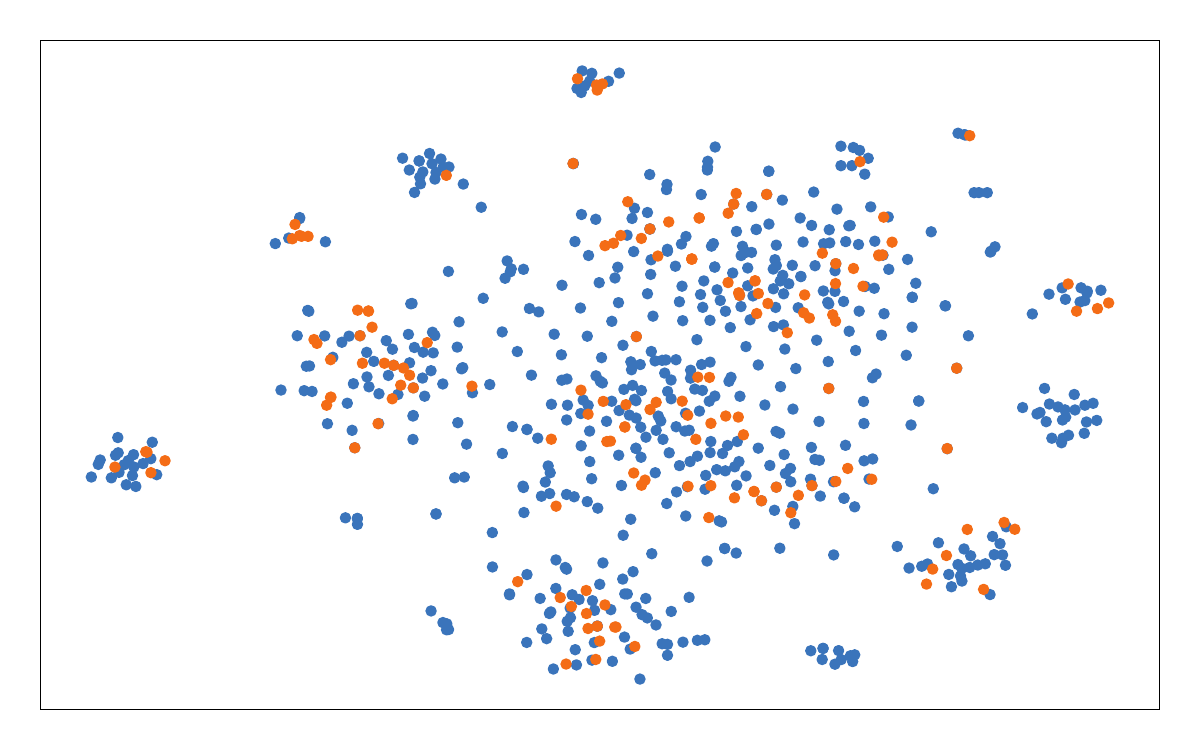} 
    \centering
    \caption{DiffGAD}
    \label{fig:diffgad_visualization}
\end{subfigure}
\begin{subfigure}[b]{.16\linewidth}
    \includegraphics[height=1.8cm]{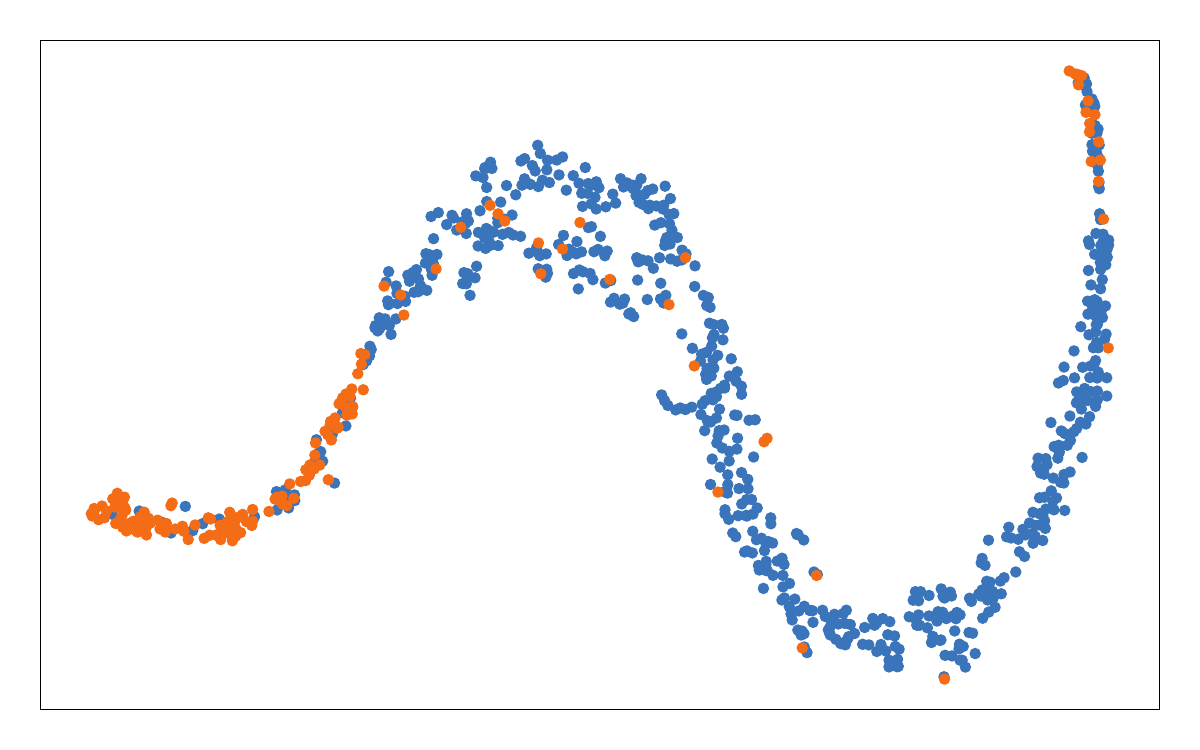} 
    \centering
    \caption{M$^2$V-UGAD}
    \label{fig:m2v_ugad_visualization}
\end{subfigure}
\caption{t-SNE embedding of various GAD methods (with GAIN as DI) and M$^2$V-UGAD on Cora under 30$\%$ missing rate.}
\label{fig:tsne}
\end{figure*}

Subsequent anomaly detectors can amplify errors or biases introduced during imputation. Contrastive learning methods like CoLA and PREM, detect anomalies based on discrepancies between nodes and their local or global neighborhoods. If imputation errors are large, these methods can inaccurately represent normal node characteristics. Conversely, strong imputation bias can homogenize reconstructed node attributes, blurring distinctions between anomalous and normal nodes. ADA-GAD can partially mitigate imputation errors through its anomaly-denoised pretraining strategy, which is evidenced by its higher performance in comparison to other baselines. However, its fine-tuning step is conducted on the fully imputed graph, making it difficult to rescue anomalies that have already been smoothed into normal patterns.

\subsection{Ablation Study}

We systematically ablate the main building blocks of M$^{2}$V-UGAD on Cora, Citeseer, Books, and Disney with 30\% missingness to gauge their individual impact (Figure~\ref{fig:ablation}). Removing either the $\LL_{\mathrm{feat}}$ imputation loss or the $\LL_{\mathrm{recon}}$ reconstruction loss produces a moderate AUROC decline ($\approx$2-4\% on most datasets), confirming that both objectives are needed to enhance imputation fidelity and maintain a well-behaved latent manifold. Eliminating an entire pathway is more damaging: dropping the feature-imputation branch lowers AUROC by roughly 3\% to 10\%, while discarding the structure-reconstruction path yields the steepest degradation among the two ($\approx$ 10-30\%). These underscore the importance of the dual-path design in preventing cross-view contamination. The most pronounced impact arises when the pseudo-anomaly generation process is disabled, slashing AUROC by roughly 10\% to 35\% across all datasets. This highlights its crucial role in countering imputation bias and sharpening the decision boundary. Collectively, these results verify that each component of M$^2$V-UGAD is indispensable for robust anomaly detection on incomplete graphs.

\subsection{Sensitivity Analysis}

We conduct a comprehensive sensitivity study to examine how four key hyper-parameters affect the performance of M$^{2}$V-UGAD under the default 30\% missingness (Figure~\ref{fig:hyper_param}). For the imputation loss weight $\alpha$, AUROC peaks when $\alpha$ lies in the corridor $0.001\leq \alpha \leq 0.01$; Setting $\alpha$ to a low number like 0.0001 may under-train the feature imputer, which is evidenced by a slight drop on Disney. Whereas setting $\alpha > 0.01$ shifts optimisation away from the latent objectives and leads to a clear decline on all datasets. For the reconstruction loss weight $\lambda$, raising $\lambda$ from 0.001 to 1 induces a monotonic yet mild downward trend, while a small dip also appears at 0.0001. These observations suggest that extremely small $\lambda$ weakens latent semantics, whereas overly large values over-regularise and blunt anomaly separation, although overall performance remains stable across roughly two orders of magnitude. In terms of the pseudo-anomaly ratio $\eta$, most datasets are almost insensitive to variations within $0.1\leq \eta \leq 0.5$; While Disney exhibits a gradual decline as $\eta$ grows. This is likely because an excessive synthetic set can dilute the decision boundary on very small graphs. Hence a moderate ratio ($\eta-0.1\sim0.3$) offers a sound trade-off. Finally, adjusting the latent-sphere radius $r$ between 6 and 8 leaves AUROC virtually unchanged, and expanding it to 10 induces only a mild decrease (below two percentage points) on Books and Disney, indicating that the Sinkhorn alignment adapts robustly provided $r$ is not extreme. Overall, the results suggest that M$^2$V-UGAD is relatively less sensitive to various parameter settings.

\begin{table*}[ht]
    \centering
    \begin{tabular}{c|c|c|c|c|c|c}
    \hline
    Dataset      & Node   & Edge   & Feature & Average Degree & Outliers & Outlier Ratio \\ \hline
    
    Cora      &  2,708    & 5,803    & 1,433 & 4.3        & 150       & 5.54\%        \\ \hline
    
    Citeseer  &  3,327    & 5,137    & 3,703 & 3.1        & 150       & 4.51\%        \\ \hline

    Books     &  1,418    & 1,847    & 21    & 2.6        & 28        & 1.97\%        \\ \hline
    
    Disney    &  124      & 167      & 28    & 2.7        & 6         & 4.84\%        \\ \hline
    
    Flickr    &  7,575    & 241,304  & 12,047& 63.7       & 450       & 5.94\%        \\ \hline   
    
    ACM       &  16,484   & 82,175   & 8,337 & 10         & 597       & 3.62\%        \\ \hline
    
    Reddit    & 10,984   & 84,008   & 64    & 15.3       & 366       & 3.33\%        \\ \hline
    
    \end{tabular}
        \caption{Statistics of datasets adopted in the experiments.}
    \label{tab:dataset-stats}
\end{table*}

\subsection{Visualization Study}

Figure~\ref{fig:tsne} shows the node embeddings from M$^2$V-UGAD alongside five GAD methods using GAIN for data imputation, on the Cora dataset with 30\% attribute and structural missingness. For the five baselines (Figure~\ref{fig:cola_visualization}-\ref{fig:diffgad_visualization}), anomalies (orange points) are clearly entangled with normal nodes (blue points), leading to blurred boundaries and difficulty in anomaly discrimination. In contrast, embeddings produced by M$^2$V-UGAD (Figure~\ref{fig:m2v_ugad_visualization}) exhibit a clear distinction, where normal nodes are arranged along a smooth manifold, while anomalous nodes are predominantly positioned on the periphery, clearly separated from the main distribution of normal nodes. This stark contrast highlights the superior ability of our proposed components, including the dual-pathway imputation, spherical latent alignment, and pseudo-anomaly generation, to preserve and emphasize anomaly signals under significant attribute and structure missingness.

%% file: conclusion_short.tex
In this paper, we present the first investigation into unsupervised graph anomaly detection under simultaneous incompleteness of node attributes and graph topology. We propose M$^2$V-UGAD, a framework combining dual-pathway imputation, Sinkhorn-driven latent regularization, and pseudo-anomaly synthesis to address feature–structure interference and imputation bias. Extensive experiments verify the effectiveness of each component and demonstrate that M$^2$V-UGAD significantly outperforms a variety of existing graph anomaly detection methods coupled with different imputation strategies across multiple benchmark datasets.

%% file: pseudo_code.tex
\begin{algorithm}
\caption{M\textsuperscript{2}V-UGAD Training and Inference}
\label{alg:mmv-ugad}
\begin{algorithmic}[1]
\STATE {\it Input:} incomplete graph $(\V,\X_{\mathrm{obs}},\A_{\mathrm{obs}},\M_X,\M_A)$; hyperparameters $\alpha,\lambda, r, r_a, r_b, \eta$; epochs $T_{\mathrm{pre}},T_{\mathrm{fine}}$
\STATE {\it Output:} anomaly scores $\{s_i\}_{i=1}^{|\V|}$ 
\STATE {\bf 1. Dual-pathway imputation}
\STATE \quad $\hat{X}\gets f_{\theta}(X_{\mathrm{obs}})$ \COMMENT{Feature path}
\STATE \quad $\hat{A} \gets A_{\mathrm{obs}} + \textsc{PPR}(A_{\mathrm{obs}})$ \COMMENT{Structure path}
\STATE {\bf 2. Pre-training}
\FOR{$t=1$ \TO $T_{\mathrm{pre}}$}
    \STATE $\mathbf{Z} \gets g_{\gamma}(\hat{\X},\hat{\A})$
    \STATE $\tilde{\X}\gets r_{\omega}(\mathbf{Z})$
    \STATE $\mathcal{L}_{\mathrm{feat}}\gets \mathrm{MSE}(\M_X\odot \hat{\X},\,\M_X\odot \X_{\mathrm{obs}})$
    \STATE $\mathcal{L}_{\mathrm{dist}}\gets \textsc{Sinkhorn}(\mathbf{Z},\mathcal{N}_{r}(0,I))$
    \STATE $\mathcal{L}_{\mathrm{recon}}\gets \mathrm{MSE}(\M_X\odot \tilde{\X},\,\M_X\odot \X_{\mathrm{obs}})$
    \STATE $\mathcal{L}_{\mathrm{pre}}\gets \alpha\mathcal{L}_{\mathrm{feat}} + \mathcal{L}_{\mathrm{dist}} + \lambda\mathcal{L}_{\mathrm{recon}}$
    \STATE Update $\{\theta,\gamma,\omega\}$ by gradient descent on $\mathcal{L}_{\mathrm{pre}}$
\ENDFOR 
\STATE {\bf 3. Pseudo-anomaly generation}
\STATE \quad $M\gets \lfloor\eta\,|\V|\rfloor$  \COMMENT{number of pseudo nodes}
\STATE \quad Sample $\{z_i^{(a)}\}_{i=1}^M$ from $\{z: r_a<\|z\|_2<r_b\}$
\STATE \quad $\tilde{\X}^{(a)}\gets r_{\omega}(\{z_i^{(a)}\})$
\STATE \quad ${\A}^{(a)}\gets \textsc{SimilarityThreshold}(\tilde{X}^{(a)},\tau_a=0.5)$
\STATE \quad $\X_{\mathrm{aug}}\gets [\,{\X_{\mathrm{obs}}};\,{\X}^{(a)}\,]$, \quad $\A_{\mathrm{aug}}\gets \begin{bmatrix}{\A_{\mathrm{obs}}}&0\\0&{\A}^{(a)}\end{bmatrix}$
\STATE {\bf 4. Fine-tuning}
\FOR{$t=1$ \TO $T_{\mathrm{fine}}$}
    \STATE $\mathbf{Z}_{\mathrm{aug}}\gets g_{\gamma}(\hat{\X}_{\mathrm{aug}},\hat{\A}_{\mathrm{aug}})$
    \STATE Split $\mathbf{Z}_{\mathrm{aug}}$ into $(\mathbf{Z}, \mathbf{Z}_{\mathrm{pseudo}})$
    \STATE $\mathcal{L}'_{\mathrm{dist}}\gets 
           \textsc{Sinkhorn}(\mathbf{Z},\mathcal{N}_{r}) + 
           \textsc{Sinkhorn}(\mathbf{Z}_{\mathrm{pseudo}},\mathcal{N}_{\{r_a<\|z\|\le r_b\}})$
    \STATE $\tilde{\X}_{\mathrm{aug}}\gets r_{\omega}(\mathbf{Z}_{\mathrm{aug}})$
    \STATE $\mathcal{L}_{\mathrm{feat}}\gets \mathrm{MSE}(\M_X\odot \hat{\X}_{\mathrm{aug}},\,\M_X\odot \X_{\mathrm{aug}})$
    \STATE $\mathcal{L}_{\mathrm{recon}}\gets \mathrm{MSE}(\M_X\odot \tilde{\X}_{\mathrm{aug}},\,\M_X\odot \X_{\mathrm{aug}})$
    \STATE $\mathcal{L}_{\mathrm{finetune}}\gets \alpha\mathcal{L}_{\mathrm{feat}} + \mathcal{L}'_{\mathrm{dist}} + \lambda\mathcal{L}_{\mathrm{recon}}$
    \STATE Update parameters by gradient descent on $\mathcal{L}_{\mathrm{fine}}$
\ENDFOR 
\STATE {\bf 5. Inference}
\FORALL{$v_i\in\V$}
    \STATE $z_i\gets g_{\gamma}(\hat{\X},\hat{\A})[i,:]$
    \STATE $s_i\gets \|z_i\|_2$
\ENDFOR
\STATE \textbf{Return} $\{s_i\}$ (ranked descending)
\end{algorithmic}
\label{Algorithm}
\end{algorithm}

%% file: aaai2026.bib
@inproceedings{yu2024towards,
  title={Towards resource-friendly, extensible and stable incomplete multi-view clustering},
  author={Yu, Shengju and Dong, Zhibin and Wang, Siwei and Wan, Xinhang and Liu, Yue and Liang, Weixuan and Zhang, Pei and Tu, Wenxuan and Liu, Xinwang},
  booktitle={Proceedings of the International Conference on Machine Learning},
   pages={57415--57440},
  year={2024}
}

@article{yu2023sparse,
  title={Sparse low-rank multi-view subspace clustering with consensus anchors and unified bipartite graph},
  author={Yu, Shengju and Liu, Suyuan and Wang, Siwei and Tang, Chang and Luo, Zhigang and Liu, Xinwang and Zhu, En},
  journal={IEEE Transactions on Neural Networks and Learning Systems},
   year={2025},
  volume={36},
  number={1},
  pages={1438-1452}
}

@inproceedings{chen2024towards,
  title={Towards cross-domain few-shot graph anomaly detection},
  author={Chen, Jiazhen and Fu, Sichao and Zhang, Zhibin and Ma, Zheng and Feng, Mingbin and Wirjanto, Tony S and Peng, Qinmu},
  booktitle={Proceedings of the IEEE International Conference on Data Mining},
  pages={51--60},
  year={2024},
  organization={IEEE}
}

@inproceedings{chen2025semi,
  title={Semi-supervised Anomaly Detection with Extremely Limited Labels in Dynamic Graphs},
  author={Chen, Jiazhen and Fu, Sichao and Ma, Zheng and Feng, Mingbin and Wirjanto, Tony S and Peng, Qinmu},
   booktitle={Proceedings of the International Conference on Database Systems for Advanced Applications},
  year={2025}
}

@inproceedings{luo2022comga,
  title={Comga: Community-aware attributed graph anomaly detection},
  author={Luo, Xuexiong and Wu, Jia and Beheshti, Amin and Yang, Jian and Zhang, Xiankun and Wang, Yuan and Xue, Shan},
  booktitle={Proceedings of the ACM International Conference on Web Search and Data Mining},
  pages={657--665},
  year={2022}
}

@inproceedings{jin2021anemone,
    title={Anemone: Graph anomaly detection with multi-scale contrastive learning},
  author={Jin, Ming and Liu, Yixin and Zheng, Yu and Chi, Lianhua and Li, Yuan-Fang and Pan, Shirui},
  booktitle={Proceedings of the ACM International Conference on Information and Knowledge Management},
  pages={3122--3126},
  year={2021}
}

@inproceedings{kaize2019dadan,
    author = {Kaize Ding and Jundong Li and Rohit Bhanushali and Huan Liu},
    title = {Deep Anomaly Detection on Attributed Networks},
    booktitle = {Proceedings of the SIAM International Conference on Data Mining},
    pages = {594-602},
    year = {2019}
}

@article{liu2022cola,
   title={Anomaly detection on attributed networks via contrastive self-supervised learning},
  author={Liu, Yixin and Li, Zhao and Pan, Shirui and Gong, Chen and Zhou, Chuan and Karypis, George},
  journal={IEEE Transactions on Neural Networks and Learning Systems},
  volume={33},
  number={6},
  pages={2378--2392},
  year={2022},
  publisher={IEEE}
}

@inproceedings{pan2023prem,
  title={PREM: A Simple Yet Effective Approach for Node-Level Graph Anomaly Detection},
  author={Pan, Junjun and Liu, Yixin and Zheng, Yizhen and Pan, Shirui},
  booktitle={Proceedings of the IEEE International Conference on Data Mining},
  pages={1253--1258},
  year={2023}
}

@inproceedings{duan2023gradate,
  title={Graph anomaly detection via multi-scale contrastive learning networks with augmented view},
  author={Duan, Jingcan and Wang, Siwei and Zhang, Pei and Zhu, En and Hu, Jingtao and Jin, Hu and Liu, Yue and Dong, Zhibin},
  booktitle={Proceedings of the AAAI Conference on Artificial Intelligence},
  volume={37},
  number={6},
  pages={7459--7467},
  year={2023}
}

@inproceedings{he2024ada,
  title={Ada-gad: Anomaly-denoised autoencoders for graph anomaly detection},
  author={He, Junwei and Xu, Qianqian and Jiang, Yangbangyan and Wang, Zitai and Huang, Qingming},
  booktitle={Proceedings of the AAAI Conference on Artificial Intelligence},
  volume={38},
  number={8},
  pages={8481--8489},
  year={2024}
}

@inproceedings{li202diffgad,
  title={DiffGAD: A Diffusion-based Unsupervised Graph Anomaly Detector}, 
  author={Jinghan Li and Yuan Gao and Jinda Lu and Junfeng Fang and Congcong Wen and Hui Lin and Xiang Wang},
  year={2025},
  booktitle={Proceedings of the International Conference on Learning Representations}
}

@article{stekhoven2011missforest,
   title={MissForest—non-parametric missing value imputation for mixed-type data},
   volume={28},
   number={1},
   journal={Bioinformatics},
   publisher={Oxford University Press (OUP)},
   author={Stekhoven, Daniel J. and Bühlmann, Peter},
   year={2011},
   pages={112–118} 
}

@InProceedings{yoon2018gain,
  title = {{GAIN}: Missing Data Imputation using Generative Adversarial Nets},
  author = {Yoon, Jinsung and Jordon, James and van der Schaar, Mihaela},
  booktitle = {Proceedings of the International Conference on Machine Learning},
  pages = {5689--5698},
  year = {2018},
  volume = {80}
}

@inproceedings{huo2023t2gnn,
  title={T2-gnn: Graph neural networks for graphs with incomplete features and structure via teacher-student distillation},
  author={Huo, Cuiying and Jin, Di and Li, Yawen and He, Dongxiao and Yang, Yu-Bin and Wu, Lingfei},
  booktitle={Proceedings of the AAAI Conference on Artificial Intelligence},
  volume={37},
  number={4},
  pages={4339--4346},
  year={2023}
}

@inproceedings{ding2019graphautoencoder,
  title={Deep anomaly detection on attributed networks},
  author={Ding, Kaize and Li, Jundong and Bhanushali, Rohit and Liu, Huan},
  booktitle={Proceedings of the SIAM International Conference on Data Mining},
  pages={594--602},
  year={2019}
}

@article{chen2020learning,
  title={Learning on attribute-missing graphs},
  author={Chen, Xu and Chen, Siheng and Yao, Jiangchao and Zheng, Huangjie and Zhang, Ya and Tsang, Ivor W},
  journal={IEEE Transactions on Pattern Analysis and Machine Intelligence},
  volume={44},
  number={2},
  pages={740--757},
  year={2020},
  publisher={IEEE}
}

@inproceedings{tu2022initializing,
  title={Initializing Then Refining: A Simple Graph Attribute Imputation Network},
  author={Tu, Wenxuan and Zhou, Sihang and Liu, Xinwang and Liu, Yue and Cai, Zhiping and Zhu, En and Zhang, Changwang and Cheng, Jieren},
  booktitle={Proceedings of the International Joint Conference on Artificial Intelligence},
  pages={3494--3500},
  year={2022}
}

@inproceedings{jiang2024incomplete,
  title={Incomplete graph learning via attribute-structure decoupled variational auto-encoder},
  author={Jiang, Xinke and Qin, Zidi and Xu, Jiarong and Ao, Xiang},
  booktitle={Proceedings of the ACM International Conference on Web Search and Data Mining},
  pages={304--312},
  year={2024}
}

@inproceedings{fu2023towards,
  title={Towards unsupervised graph completion learning on graphs with features and structure missing},
  author={Fu, Sichao and Peng, Qinmu and He, Yang and Du, Baokun and You, Xinge},
  booktitle={Proceedings of the IEEE International Conference on Data Mining},
  pages={1019--1024},
  year={2023}
}

@inproceedings{zemicheal2019anomaly,
  title={Anomaly detection in the presence of missing values for weather data quality control},
  author={Zemicheal, Tadesse and Dietterich, Thomas G},
  booktitle={Proceedings of the ACM SIGCAS Conference on Computing and Sustainable Societies},
  pages={65--73},
  year={2019}
}

@inproceedings{xiao2024unsupervised,
    title={Unsupervised Anomaly Detection in The Presence of Missing Values},
    author={Feng Xiao and Jicong Fan},
    booktitle={Proceedings of the Annual Conference on Neural Information Processing Systems},
    year={2024}
}

@article{park2019survey,
  title={A survey on personalized PageRank computation algorithms},
  author={Park, Sungchan and Lee, Wonseok and Choe, Byeongseo and Lee, Sang-Goo},
  journal={IEEE Access},
  volume={7},
  pages={163049--163062},
  year={2019},
  publisher={IEEE}
}

@article{ma2021comprehensive,
  title={A comprehensive survey on graph anomaly detection with deep learning},
  author={Ma, Xiaoxiao and Wu, Jia and Xue, Shan and Yang, Jian and Zhou, Chuan and Sheng, Quan Z and Xiong, Hui and Akoglu, Leman},
  journal={IEEE Transactions on Knowledge and Data Engineering},
  volume={35},
  number={12},
  pages={12012--12038},
  year={2021},
  publisher={IEEE}
}

@inproceedings{chaudhary2019anomaly,
  title={Anomaly detection using graph neural networks},
  author={Chaudhary, Anshika and Mittal, Himangi and Arora, Anuja},
  booktitle={Proceedings of the International Conference on Machine Learning, Big Data, Cloud and Parallel Computing},
  pages={346--350},
  year={2019}
}

@inproceedings{fakhraei2015collective,
  title={Collective spammer detection in evolving multi-relational social networks},
  author={Fakhraei, Shobeir and Foulds, James and Shashanka, Madhusudana and Getoor, Lise},
  booktitle={Proceedings of the ACM SIGKDD International Conference on Knowledge Discovery and Data Mining},
  pages={1769--1778},
  year={2015}
}

@inproceedings{zhang2020gcn,
  title={Gcn-based user representation learning for unifying robust recommendation and fraudster detection},
  author={Zhang, Shijie and Yin, Hongzhi and Chen, Tong and Hung, Quoc Viet Nguyen and Huang, Zi and Cui, Lizhen},
  booktitle={Proceedings of the International ACM SIGIR Conference on Research and Development in Information Retrieval},
  pages={689--698},
  year={2020}
}

@inproceedings{huang2022dgraph,
  title={Dgraph: A large-scale financial dataset for graph anomaly detection},
  author={Huang, Xuanwen and Yang, Yang and Wang, Yang and Wang, Chunping and Zhang, Zhisheng and Xu, Jiarong and Chen, Lei and Vazirgiannis, Michalis},
  booktitle={Advances in Neural Information Processing Systems},
  volume={35},
  pages={22765--22777},
  year={2022}
}

@article{zhang2022efraudcom,
  title={efraudcom: An e-commerce fraud detection system via competitive graph neural networks},
  author={Zhang, Ge and Li, Zhao and Huang, Jiaming and Wu, Jia and Zhou, Chuan and Yang, Jian and Gao, Jianliang},
  journal={ACM Transactions on Information Systems},
  volume={40},
  number={3},
  pages={1--29},
  year={2022},
  publisher={ACM New York, NY}
}

@article{yuan2024mds,
  title={MDS-GNN: A Mutual Dual-Stream Graph Neural Network on Graphs with Incomplete Features and Structure},
  author={Yuan, Peng and Tang, Peng},
  journal={arXiv preprint arXiv:2408.04845},
  year={2024}
}

@article{xia2025incomplete,
  title={Incomplete graph learning: A comprehensive survey},
  author={Xia, Riting and Liu, Huibo and Li, Anchen and Liu, Xueyan and Zhang, Yan and Zhang, Chunxu and Yang, Bo},
  journal={Neural Networks},
  pages={107682},
  year={2025},
  publisher={Elsevier}
}

@article{jiang2020incomplete,
  title={Incomplete graph representation and learning via partial graph neural networks},
  author={Jiang, Bo and Zhang, Ziyan},
  journal={arXiv preprint arXiv:2003.10130},
  year={2020}
}

@inproceedings{ruff2018deep,
  title={Deep one-class classification},
  author={Ruff, Lukas and Vandermeulen, Robert and Goernitz, Nico and Deecke, Lucas and Siddiqui, Shoaib Ahmed and Binder, Alexander and M{\"u}ller, Emmanuel and Kloft, Marius},
  booktitle={Proceedings of the International Conference on Machine Learning},
  pages={4393--4402},
  year={2018}
}

@inproceedings{kipf2016variational,
  title={Variational Graph Auto-Encoders},
  author={Kipf, Thomas N and Welling, Max},
  booktitle={Advances in Neural Information Processing Systems Workshop},
  year={2016}
}

@inproceedings{yang2016revisiting,
  title={Revisiting semi-supervised learning with graph embeddings},
  author={Yang, Zhilin and Cohen, William and Salakhudinov, Ruslan},
  booktitle={Proceedings of the International Conference on Machine Learning},
  pages={40--48},
  year={2016}
}

@inproceedings{zeng2019graphsaint,
  title={Graphsaint: Graph sampling based inductive learning method},
  author={Zeng, Hanqing and Zhou, Hongkuan and Srivastava, Ajitesh and Kannan, Rajgopal and Prasanna, Viktor},
  booktitle={Proceeding of the International Conference on Learning Representations}, 
  year={2020}
}

@inproceedings{kumar2019reddit,
  author = {Kumar, Srijan and Zhang, Xikun and Leskovec, Jure},
  booktitle = {Proceeding of the ACM SIGKDD International Conference on Knowledge Discovery and Data Mining},
  pages = {1269-1278},
  title = {Predicting Dynamic Embedding Trajectory in Temporal Interaction Networks},
  year = {2019}
}

@inproceedings{muller2013disney,
  author = {Müller, Emmanuel and Sánchez, Patricia Iglesias and Mülle, Yvonne and Böhm, Klemens},
  booktitle = {Proceeding of the IEEE International Conference on Data Engineering Workshops},
  pages = {216-222},
  title = {Ranking outlier nodes in subspaces of attributed graphs},
  year = {2013}
}

@inproceedings{sanchez2013books,
  author = {Sánchez, Patricia Iglesias and Müller, Emmanuel and Laforet, Fabian and Keller, Fabian and Böhm, Klemens},
  booktitle = {Proceeding of the IEEE International Conference on Data Mining},
  pages = {647-656},
  title = {Statistical Selection of Congruent Subspaces for Mining Attributed Graphs},
  year = {2013}
}

@inproceedings{tang2008acm,
  author = {Tang, Jie and Zhang, Jing and Yao, Limin and Li, Juanzi and Zhang, Li and Su, Zhong},
  booktitle = {Proceeding of the ACM SIGKDD International Conference on Knowledge Discovery and Data Mining},
  numpages = {9},
  pages = {990--998},
  title = {ArnetMiner: extraction and mining of academic social networks},
  year = {2008}
}
